\documentclass[10pt,journal,compsoc]{IEEEtran}
\ifCLASSOPTIONcompsoc
  \usepackage[nocompress]{cite}
\else
  \usepackage{cite}
\fi
\usepackage{graphicx}
\usepackage{epstopdf}
\usepackage{multirow}
\usepackage{amsmath}
\usepackage{amssymb}
\usepackage[normalem]{ulem}
\useunder{\uline}{\ul}{}
\usepackage{booktabs}
\usepackage{units}
\usepackage{color}
\ifCLASSINFOpdf
\else
\fi
\usepackage{url}

\begin{document}
%
\title{Learning Gait Representation from Massive Unlabelled Walking Videos: A Benchmark}
%
%
%
%

\author{Chao~Fan,
        Saihui~Hou,
        Jilong~Wang, 
        Yongzhen~Huang, 
        and~Shiqi~Yu
\IEEEcompsocitemizethanks{\IEEEcompsocthanksitem 
Chao Fan and Shiqi Yu are with Department of Computer Science and Engineering, Southern University of Science and Technology, 
Shenzhen 518055, China. E-mail: 12131100@mail.sustech.edu.cn and yusq@sustech.edu.cn. 
\IEEEcompsocthanksitem 
Saihui Hou and Yongzhen Huang are with School of Artificial Intelligence, Beijing Normal University and also with Watrix Technology Limited Co. Ltd., China. E-mail: \{huangyongzhen, housaihui\}@bnu.edu.cn.
\IEEEcompsocthanksitem 
Jilong Wang is with the University of Science and Technology of China and also with Center for Research on Intelligent Perception and Computing (CRIPAC), National Laboratory of Pattern Recognition (NLPR), Institute of Automation, Chinese Academy of Sciences (CASIA). Email: jilongw@mail.ustc.edu.cn.
\IEEEcompsocthanksitem 
Corresponding author: Shiqi Yu.
\protect\\
}
\thanks{Manuscript received June 08, 2022.}
}

%
%

\markboth{IEEE Transactions on Pattern Analysis and Machine Intelligence,~Vol.~XX, No.~YY, Month~2022}%
{Fan \MakeLowercase{\textit{et al.}}: A Large-Scale Self-Supervised Benchmark for Gait Recognition with Contrastive Learning}
%



\IEEEtitleabstractindextext{%
\begin{abstract}
Gait depicts individuals' unique and distinguishing walking patterns 
and has become one of the most promising biometric features for human identification.
As a fine-grained recognition task, 
gait recognition is easily affected by many factors and usually requires a large amount of completely annotated data that is costly and insatiable.
This paper proposes a large-scale self-supervised benchmark for gait recognition with contrastive learning, 
aiming to learn the general gait representation from massive unlabelled walking videos for practical applications via offering informative walking priors and diverse real-world variations.
Specifically, we collect a large-scale unlabelled gait dataset GaitLU-1M consisting of 1.02M walking sequences and propose a conceptually simple yet empirically powerful baseline model GaitSSB.
Experimentally, 
we evaluate the pre-trained model on four widely-used gait benchmarks, 
CASIA-B, OU-MVLP, GREW and Gait3D with or without transfer learning.
The unsupervised results are comparable to or even better than the early model-based and GEI-based methods.
After transfer learning, 
GaitSSB outperforms existing methods by a large margin in most cases\textcolor{black}{, and also showcases the superior generalization capacity. Further experiments indicate that the pre-training can save about 50\% and 80\% annotation costs of GREW and Gait3D}. 
Theoretically, 
we discuss the critical issues for gait-specific contrastive framework and present some insights for further study.
As far as we know, GaitLU-1M is the first large-scale unlabelled gait dataset, and GaitSSB is the first method that achieves remarkable unsupervised results on the aforementioned benchmarks.
\textcolor{black}{The source code of GaitSSB and anonymous data of GaitLU-1M is available at \url{https://github.com/ShiqiYu/OpenGait}}. 
\end{abstract}

\begin{IEEEkeywords}
Gait Recognition, Self-Supervised, Contrastive Learning, GaitSSB, GaitLU-1M.
\end{IEEEkeywords}}

\maketitle

\IEEEdisplaynontitleabstractindextext

%
\IEEEpeerreviewmaketitle

\IEEEraisesectionheading{\section{Introduction}\label{sec:introduction}}

%
%
%
%

 
\IEEEPARstart{G}{ait} recognition aims to capture human walking patterns by computer vision technologies for individual identification.
Compared with other vision-based biometrics, \textit{e.g.}, face, fingerprint and iris, 
gait is hard to disguise and can be easily depicted at a long distance in non-intrusive ways without cooperative subjects.
These advantages make gait recognition particularly suitable for various security applications, 
such as suspect tracking, crime prevention, identity verification, and so on\cite{Wu2016}.

\begin{figure*}[htbp]
\centering
\includegraphics[height=4.0cm]{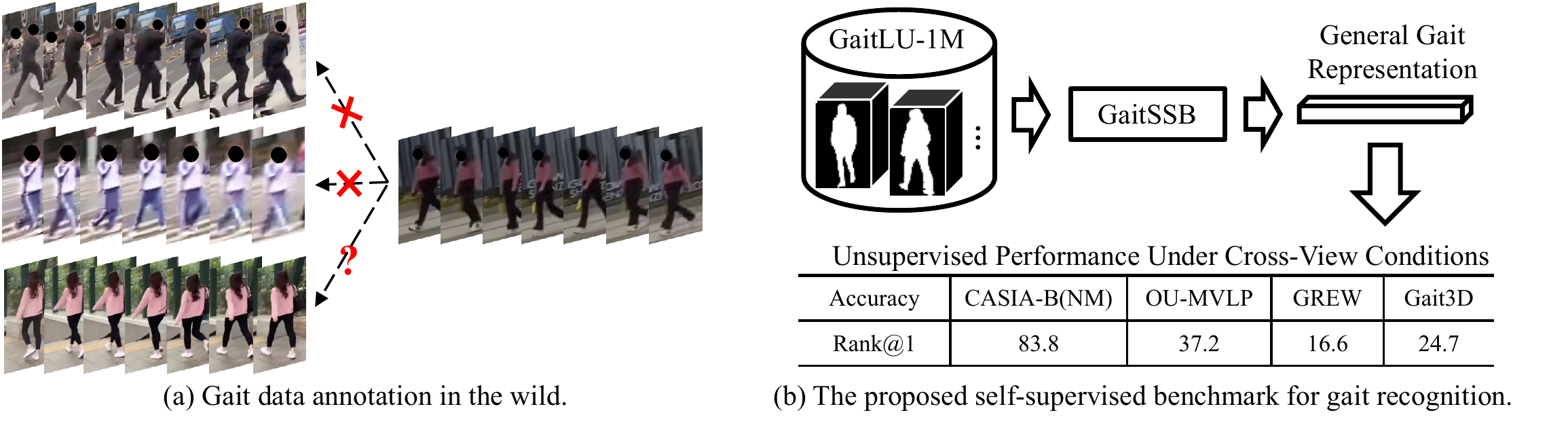}
\caption{
(a) It is generally hard for an annotator to identify individuals according to their walking patterns without the help of extra attributes in practical scenarios.
(b) Our prospective and comprehensive self-supervised benchmark with contrastive learning for gait recognition, 
including a large-scale unlabelled gait dataset GaitLU-1M and a conceptually simple yet empirically powerful baseline model GaitSSB.}
\label{fig1}
\vspace{-1.0em}
\end{figure*}

Recently, with the rising of deep learning\cite{lecun2015deep}, the gait recognition community has continuously attracted increasing interest from both academic and industrial fields since the promising performance achieved on multiple popular gait benchmarks, such as CAISA-B\cite{Yu2006} and OU-MVLP\cite{Takemura2018}.
However, evidence from \cite{zhu2021gait} and \cite{zhang2022realgait} indicate that existing methods\cite{Chao2019, fan2020gaitpart, gaitgl} are still far away from practical applications.
One of the primary reasons is that almost all the gait datasets are collected by fixed cameras under a fully-controlled environment, \textit{e.g.}, indoor laboratory or specific areas in the wild, and therefore struggle to simulate the fully-unconstrained challenges, 
such as diverse camera view and height, complex occlusion and background, various carrying and dressing, unpredictable illumination and weather, \textit{et al}.
One straightforward solution is collecting a large amount of real gait data to fill the massive gap between theory and practice.
However, large-scale gait data annotation in the wild is particularly expensive. 
Specifically, 
unlike the face and pedestrian that can be easily identified by just one image, 
the gait characteristics are generally represented as the combination of static appearance and dynamic movement patterns inherent in the walking videos, 
thus requiring much time and effort for annotations.
More importantly, as shown in Fig.~\ref{fig1}(a), if there are no other apparent soft-biometric features or auxiliary information, such as gender, age, dress, location, background, and so on, it would be almost impossible for an annotator to identify individuals from massive unlabelled videos.

Nowadays, the fact becomes that, despite gait data with manual labels being costly, 
there are massive unlabelled walking videos on the web and apps, such as the prevailing YouTube and TikTok.
Furthermore, pre-training on unlabelled data by self-supervised approaches has been believed to effectively explore the data-specific prior knowledge so as to alleviate the 'out-of-distribution' problem.
These advantages make self-supervised methods highly successful in both natural language processing\cite{radford2018improving, bert} and computer vision fields in recent years.
Specifically, 
the impressive BERT\cite{bert} can be fine-tuned with just one additional output layer to achieve state-of-the-art for eleven language tasks.
For visual tasks, 
reducing the distance between representations of two different augmented views of the same image (positive pairs), 
and meanwhile increasing the distance between representations of different images (negative pairs)\cite{grill2020bootstrap}, 
\textit{i.e.}, the contrastive learning framework has been steady progress in self-supervised representation learning, 
with encouraging results on multiple visual tasks\cite{he2020momentum, chen2020simple, grill2020bootstrap, simsiam}.
However, existing gait recognition methods 
are still immersed in relatively small-scale supervised datasets.
More importantly, we thus lack a well pre-trained model providing knowledgeable walking priors obtained by dealing with substantial real-world factors.

To address these issues, we develop a pioneering and comprehensive self-supervised benchmark to learn the general gait representation from massive unlabelled walking videos.
Firstly, a new dataset is required. 
Hence we collect a \textbf{L}arge-scale \textbf{U}nlabelled gait dataset and name it as \textbf{GaitLU-1M}.
Inspired by \cite{abu2016youtube, xu2018youtube, fu2021}, 
GaitLU-1M is extracted from 1.6K publicly available videos and consists of 1.02 million gait sequences, whose training set is about 10 times larger than that of the largest gait datasets, \textit{e.g.}, OU-MVLP\cite{Takemura2018} and GREW\cite{zhu2021gait}.
Moreover, the collected videos are filmed by different cameras worldwide, covering a wide range of capturing scenes, individual attributes, photography conditions, \textit{etc}.

In addition, we introduce contrastive learning, one of the most popular visual self-supervised methods, to gait recognition and point out two gait-specific issues.
The first is about \textit{input modality}.
Human gait patterns captured by walking videos involve dynamic motion information along the temporal dimension, which differs a lot from a single RGB image,
indicating that some crucial and well-explored data augmentation strategies for common contrastive framework\cite{grill2020bootstrap} are no longer applicable to gait recognition.
In this paper, 
we propose a silhouette augmentation operator $\pi(\cdot)$ and find that the gait-specific augmentation strategy is vital for contrastive gait recognition.
Another issue is about the \textit{task granularity}.
The phenomenon that two different images from semantically near categories own relatively lower visual distance and vice versa has been considered helpful for contrastive framework\cite{Wu_2018_CVPR}.
However, gait recognition is a fine-grained identification task aiming to identify individuals, 
whose one of the most prominent features is that intra-class distances are often larger than inter-class, 
especially for cross-view cases.
Through experimental and theoretical analysis, 
we find that the negative samples play an essential role in learning view-invariant gait characteristics.
Taking the above insights together, 
we propose a conceptually simple and empirically powerful model, \textbf{GaitSSB}, 
to build a comprehensive \textbf{S}elf-\textbf{S}upervised \textbf{B}enchmark for gait.
Overall, this paper makes the following contributions.
\begin{itemize}
\item We successfully leverage the massive unlabelled walking videos to become beneficial for gait recognition.
Specifically, we propose a pioneering and comprehensive self-supervised benchmark with contrastive learning, 
including a large-scale unlabelled gait dataset GaitLU-1M and a conceptually simple yet empirically powerful model GaitSSB.
\item As far as we know, GaitSSB is the first method achieving noteworthy unsupervised performance on four of the most popular supervised gait benchmarks, \textit{i.e.}, $83.8\%$, $35.0\%$, $16.6\%$ and $24.7\%$ rank-1 average accuracy under cross-view conditions on CASIA-B (normal walking), OU-MVLP, GREW and Gait3D, respectively.
\item With transfer learning, GaitSSB outperforms existing methods by a large margin on two practical gait datasets (GREW and Gait3D).
On two indoor gait datasets (OU-MVLP and CASIA-B), GaitSSB can also achieve state-of-the-art or fully comparable performances. 
\textcolor{black}{Moreover, under the challenging cross-domain settings, GaitSSB further exhibits its superior capacity of generalization. 
Experiments also show that the pre-training can save about 50\% and 80\% annotation costs of GREW and Gait3D.
}
\item We discuss the crucial issues for introducing the contrastive framework into gait recognition from both experimental and theoretical perspectives, convincingly boosting the further explorations of learning the general gait representation for practical gait recognition applications.
\end{itemize}

In the remaining part of this paper, after introducing more related works on gait recognition and contrastive learning in Sec.~\ref{sec2}, 
we will present the process of building GaitLU-1M and show the details of GaitSSB in Sec.~\ref{sec3}.
Experimental results, theoretical discussions and brief conclusions are respectively given in Sec.~\ref{sec4}, Sec.~\ref{sec5} and Sec.~\ref{sec6}.
In Sec.~\ref{sec7}, we state the concerns about privacy protection and biometrics abuse.

\section{Related Works}
\label{sec2}
\subsection{Gait Recognition}
Almost all the existing gait recognition methods are based on supervised learning with labeled datasets, 
and can be generally grouped into two distinct categories, involving the model-based and appearance-based methods.

\textcolor{black}{
\textit{Model-based} 
methods\cite{kusakunniran2009multiple,wang2006gait,bodor2009view,ariyanto2011model,liao2020model} aim to construct the underlying structure of the human body first and then extract gait features for individual identification. 
Thanks to its compactness, the skeleton data is a kind of human body model usually used for gait representation learning~\cite{liao2020model, teepe2021gaitgraph, teepe2022towards}.
In the latest literature, the human mesh that preserves rich 3D characteristics is becoming increasingly popular~\cite{li2022multi, li2021end, li2020end, xu2023occlusion}. 
Due to the nature of directly representing the dynamic changes of body parts and being robust to real-world factors like dressing and carrying, these body models that exclude the visual clues by merely maintaining the structural information are considered a clean representation of gait in theory. 
In practice, lacking the description of body shape usually makes the performance of skeleton-based methods relatively unsatisfactory~\cite{liao2020model, teepe2021gaitgraph, teepe2022towards}. 
On the other hand, the mesh-based methods representing human body surfaces have achieved more competitive recognition accuracy~\cite{li2022multi, li2021end, li2020end, xu2023occlusion}. 
}

\textcolor{black}{
\textit{Appearance-based} methods usually extract the gait features from silhouettes directly and thus benefit from the informative shape features}. 
According to the input class, these methods can be roughly divided into two categories: template-based\cite{kusakunniran2013new, makihara2006gait, han2005individual, wang2011human, Li_2020_CVPR, Zhang_2019_CVPR} and sequence-based\cite{Chao2019, fan2020gaitpart, hou2020gait, chi2020eccv, gaitgl, 3DLocal, CSTL, Zhang_2021_CVPR, zhang2020learning}, 
where the former usually compresses the gait sequence into one template along the temporal dimension while the latter tends to regard the walking video as an ordered sequence\cite{fan2020gaitpart, gaitgl, zhang2020learning} or an unordered set\cite{Chao2019}.
With the rising of deep learning, this kind of method mainly focuses on the design of backbone structures (\textit{e.g.}, set block proposed by \cite{Chao2019}, focal convolution proposed by \cite{fan2020gaitpart}, lateral connection proposed by \cite{hou2020gait}, and 3D global-local convolution proposed by \cite{gaitgl, 3DLocal}.), 
the neck of network (\textit{e.g.}, horizontal pyramid pooling introduced by \cite{Chao2019}, micro-motion module proposed by \cite{fan2020gaitpart},  generalized-mean pooling introduced by \cite{gaitgl}), 
and the head of network (\textit{e.g.}, triplet loss introduced by \cite{Chao2019} and quintuplet loss proposed by \cite{Zhang_2019_CVPR}).

Additionally, there raise a series of end-to-end based gait recognition methods directly taking RGB images as input\cite{song2019gaitnet, zhang2020learning, li2020end, liang2022gaitedge}.
They usually eliminate the gait-unrelated features by carefully designing the intermediate modality, such as float mask\cite{song2019gaitnet}, SMPL model\cite{li2020end} and synthetic silhouettes\cite{liang2022gaitedge}.
Despite achieving impressive performances, they have not attracted enough attention due to the lack of public RGB-based gait datasets.

Besides, like other vision tasks, 
the development of gait datasets has greatly promoted gait recognition research.
Specifically, CASIA-B\cite{Yu2006} and OU-MVLP\cite{Takemura2018} are two of the most widely-used gait datasets.
They were proposed in 2006 and 2018, respectively, and captured by requiring the subjects to walk around the laboratory, differing significantly from the real-world scenarios.
With facing more practical applications, 
GREW\cite{zhu2021gait} and Gait3D\cite{zheng2022gait3d} are collected in the wild in 2021 and 2022 respectively.
Though being collected from outdoor areas, 
they are still inferior in both scale and diversity compared to other biometrics recognition datasets, \textit{e.g.}, MegaFace\cite{kemelmacher2016megaface} for face recognition.

\subsection{Contrastive Methods for Self-Supervised Learning}
The success of self-supervised learning is mainly owing to its ability to explore valuable prior knowledge from massive unlabelled data.
The contrastive learning\cite{hadsell2006dimensionality}, whose core idea is to attract two different augmented views\footnote{A view generally means a glance at the augmented image in contrastive learning, but indicates the camera angle in gait recognition.} of the same image and simultaneously enlarge the distance between representations of two different images, has been popularized for self-supervised representation learning due to its state-of-the-art performance and excellent capacity of generalization.

In practice, current contrastive learning methods mainly involve several aspects of designs, 
\textit{i.e.}, data augmentation to form positive pairs, various encoders to embed input images, and pretext task as a training goal for learning good data representations.
Specifically, 
Chen \textit{et al.} demonstrate that the composition of data augmentations plays a critical role in contrastive learning,
and most of the follow-on works agree with this conclusion\cite{chen2020improved, grill2020bootstrap, simsiam, caron2020unsupervised}.
For the encoder type, 
with mostly taking ResNet\cite{he2016deep} or ViT\cite{dosovitskiy2020image} as backbone, 
some works\cite{simsiam, chen2020simple} employ the widely-used siamese networks\cite{bromley1993signature} to compare entities, 
and others tend to maintain an extra momentum memory bank to remain massive negative samples\cite{grill2020bootstrap, he2020momentum, chen2020improved} or clustering centers\cite{caron2020unsupervised} to build the large and informative training batch.

Pretext tasks, where \emph{pretext} implies that the solved task is not of genuine interest but is solved only for the true purpose of learning a good representation\cite{simsiam}, can  distinguish the categories of self-supervised methods clearly,
\textit{e.g.}, pixel-level image reconstruction for denoising autoencoder\cite{vincent2008extracting}, 
patch-level image reconstruction for masked autoencoder\cite{he2021masked}, 
and language probability model for word vector embedding\cite{mikolov2013efficient}.
There are many pretext tasks usually employed in contrastive framework, \textit{e.g.}, context auto-encoding in \cite{oord2018representation} and contrastive multiview coding in \cite{tian2020contrastive}. 

\subsection{Gait Recognition with Contrastive Learning}
Before the boost of deep learning, 
there are many classical works extracting the gait representation via unsupervised approaches, 
such as Principal Components Analysis (PCA)\cite{murase1996moving} and Canonical Analysis\cite{huang1999recognising}.
The main focus of them is to capture the valuable features used as the input of the downstream gait pattern classifier, 
rather than directly pre-train the model on massive unlabelled walking videos. 
In the latest literature, 
some works~\cite{rao2021self, liu2021selfgait} have mentioned the idea of contrastive learning 
but have not performed significantly better than other progressive methods.
Besides, instead of the large-scale practical unlabelled datasets, these works \cite{rao2021self, liu2021selfgait} are developed on the relatively small-scale and fully-constrained CAISA-B\cite{Yu2006} or OU-MVLP\cite{Takemura2018}.
Hence, they are believed to deviate from the primary intention of pre-training, that is to explore the prior knowledge from massive unlabelled walking videos for the excellent capacity of representation and generalization.

\begin{figure*}[htbp]
\centering
\includegraphics[height=4.3cm]{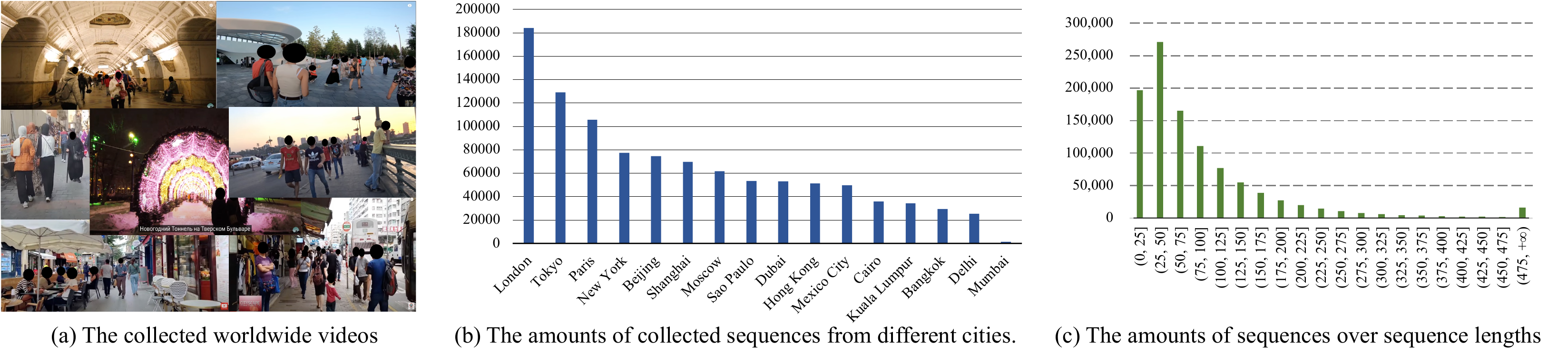}
\caption{(a) shows the example of collected public videos worldwide by different devices. (b) and (c) present the statistics of GaitLU-1M.}
\label{fig2}
\vspace{-1.0em}
\end{figure*}

\begin{table*}[htbp]
\centering
\caption{
The comparison between widely-used gait datasets with GaitLU-1M. \#Id, \#Seq and \#Cam refer to the count of identities, sequences and cameras, respectively. NM, BG and CL indicate normal, carrying and dressing factors respectively. $^\asymp$ indicates the approximate number.}
\label{tab1}
\begin{tabular}{ccccccccc}
\toprule
Dataset   & Source   & Labelled & \#Id      & \#Seq     & \#Cam & Environment     & Viewpoint    & Variations \\ \midrule 
CASIA-B   & ICPR2006 & Yes      & 124       & 13,640    & 11    & laboratory      & fixed   & NM, BG, CL \\
OU-MVLP    & CVA2018  & Yes      & 10,307     & 288,596   & 14    & laboratory      & fixed   & NM         \\
GREW      & ICCV2021 & Yes      & 26,345     & 128,671   & 882   & particular area & free    & real world \\
Gait3D    & CVPR2022 & Yes      & 4,000      & 25,309    & 39    & supermarket     & free    & real world \\
GaitLU-1M & ours        & No       & $1,035,309^\asymp$ & 1,035,309 & 1,379  & open world      & free    & real world \\ \bottomrule 
\end{tabular}
\vspace{-1.0em}
\end{table*}

In this paper, 
we constructively point out two gait-specific issues distinct from canonical contrastive frameworks,
involving input modality (sequence of silhouettes \textit{v.s.} RGB image) and task granularity (fine-grained recognition \textit{v.s.} common object classification).
These new issues make it uncertain whether contrastive learning would still work well for gait recognition.
Through careful designs, 
we build a pioneer and comprehensive benchmark for contrastive gait recognition, 
including a large-scale unlabelled gait dataset GaitLU-1M
and a conceptually simple yet empirically powerful model GaitSSB.

\subsection{Contrastive Learning on Related Tasks}
Here we discuss two related tasks, 
\textit{i.e.}, person re-identification and spatiotemporal representation learning.

\textit{Person Re-ID.} Similar to gait recognition, 
person Re-ID aims to find the target person presented in CCTV cameras.
Recently, 
SpCL\cite{ge2020self} fine-tunes the network using pseudo labels generated from reliable clustering results trained with contrastive loss.
Fu \textit{et al.}\cite{fu2021unsupervised} present a large-scale unlabelled person re-identification dataset extracted from public videos and discover some unique factors specific to Re-ID task, including that color distortion is harmful, Random Erasing\cite{zhong2020random} is needed, and a proper temperature parameter for contrastive loss is important.

\textit{Spatiotemporal representation learning.} 
Feichtenhofer \textit{et al.}\cite{feichtenhofer2021large} present a large-scale study on contrastive spatiotemporal representation learning and find that temporally-persistent features in the same video work surprisingly well across different contrastive frameworks, pre-training datasets, downstream datasets and encoder architectures.

Gait recognition differs from the above-related tasks in both the input modality, 
\textit{i.e.}, sequence of silhouettes \textit{v.s.} a single or sequence of RGB images, 
and task granularity, \textit{i.e.}, fine-grained biometric recognition \textit{v.s.} coarse-grained human appearance or action recognition. 
But this paper gets some consistent conclusions from them, 
\textit{e.g.}, gait-specific data augmentation strategies are necessary, and consistencies features along temporal dimensions are crucial.

\section{Proposed Benchmark}
\label{sec3}
\subsection{GaitLU-1M}
Collecting a large-scale unlabelled gait dataset depicting various practical factors is a prerequisite for constructing the self-supervised gait benchmark.
However, existing gait datasets either capture the walking videos in an indoor laboratory, \textit{e.g.}, CASIA-B\cite{Yu2006} and OU-MVLP\cite{Takemura2018}, or in particular areas in the wild, \textit{e.g.}, GREW\cite{zhu2021gait} and Gait3D\cite{zheng2022gait3d}, 
struggling from simulating the fully open real-world challenges.
Moreover, the scale of gait datasets is also far less than that of related fields, 
\textit{e.g.}, MegaFace\cite{kemelmacher2016megaface} that consists of millions of images used for face recognition.

In this work, we build a \textbf{L}arge-scale \textbf{U}nlabelled gait dataset, \textbf{GaitLU-1M}.
It consists of over one million walking sequences averaging 92 frames, namely about 92M silhouettes, 
76 times larger than the number of images in the famous ImageNet-1k (1.2M images)\cite{deng2009imagenet}.
Inspired by \cite{abu2016youtube, xu2018youtube} and \cite{fu2021}, 
GaitLU-1M is extracted from public videos shot around the world, 
making it cover a wide range of variations,  
involving the capturing scenes (streets, subway and airport, \textit{etc.}), individual attributes (gender, age and race, \textit{etc.}), photography conditions (camera view, resolution, illumination and weather, \textit{etc.}), and so on.
This great diversity and scale offer an excellent chance to learn general gait representation in a self-supervised manner.

\begin{figure*}[htbp]
\centering
\includegraphics[height=3.2cm]{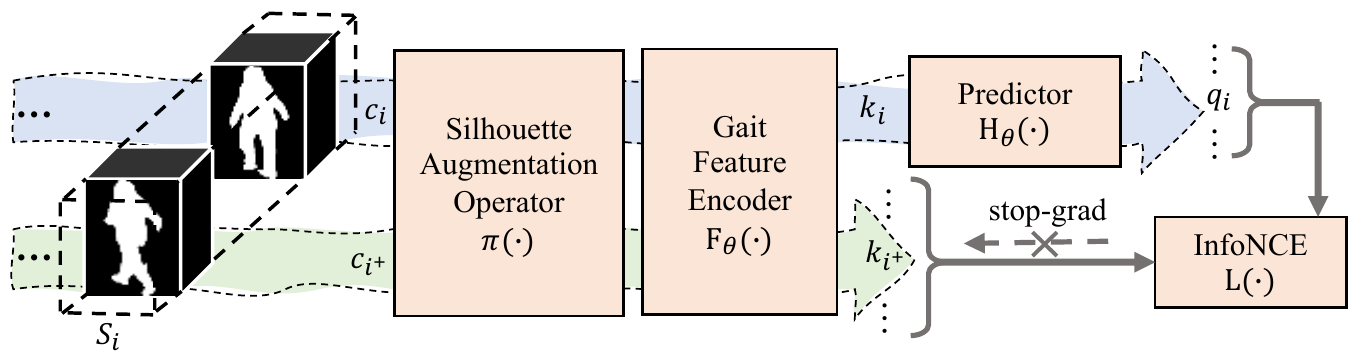}
\caption{The overall of GaitSSB. $c^+_i$ and $c_i$ denotes two non-overlap clips of the gait sequence $S_i$ where $i\in\left \{1, ..., n  \right \}$.
And $\pi(\cdot )$, $\textrm{F}_\theta (\cdot)$, $\textrm{H}_\theta (\cdot)$ and $\textup{\textrm{L}}(\cdot )$ indicates the silhouette augmentation operator, gait feature encoder, predictor and InfoNCE loss, respectively.}
\label{fig3}
\vspace{-1.0em}
\end{figure*}
\subsubsection{Data Collection and Processing}
By using queries like "city name + walking", we choose 16 big cities around the world. There are 3 American cities (New York, Sao Paulo and Mexico City), 3 European cities (London, Paris and Moscow), 9 Asian cities (Beijing, Hong Kong, Shanghai, Tokyo, Bangkok, Delhi, Mumbai, Kuala Lumpur and Dubai) and 1 African city (Cairo).
The top 100 searched videos will be collected for each city, 
that is, we crawled about 1.6K videos in total. 
Besides, the videos appearing with no person in the first 2 minutes are skipped for efficiency, 
and 221 videos are ignored in this way.

Then the pedestrian tracking algorithm, BYTETracker\cite{zhang2021bytetrack} is performed to provide initial sequences of human bounding boxes.
Since most videos are filmed in crowded public places, 
such as streets, squares, airports, and subway stations, 
we obtain over 3M raw walking sequences even if the average duration of collected videos is just a half hour.
However, we find a significant portion of the gained boxes suffers from severe occlusion because of being in crowds and not looking like the human appearance.
Therefore, we discard those small bounding boxes in each image if they overlap others over an IOU-measured threshold for data cleaning.
After performing the pedestrian segmentation\cite{yuan2020object,liu2021paddleseg} frame by frame, 
we get 1,035,309 walking sequences containing 92,627,263 silhouettes. 
Some example videos are given in Fig.~\ref{fig2} (a),
showing that GaitLU-1M has pretty diverse variations, such as camera angle, background and individual race.
\textit{
To protect personal privacy, 
the obtained gait videos are anonymous with random index and exhibited as the sequences of binary silhouette \textcolor{black}{and body skeleton.}
\textcolor{black}{ The pre-processing is based on automatic algorithms, meaning that nobody has manually processed personal visual data.
The raw data, including website links, RGB videos and other intermediate results, have been deleted. }
}

\subsubsection{Statistics and Comparison}
Statistics about GaitLU-1M are illustrated in Fig.~\ref{fig2} (b) and (c). 
Among them, Fig.~\ref{fig2} (b) shows the amounts of sequences collected from different cities, and more sequences can be collected from the cities with relatively large populations. 
Fig.~\ref{fig2} (c) tells over 80\% of gait sequences contain over 25 frames, namely including about one gait cycle at least.

Compared with existing gait datasets, such as CASIA-B\cite{Yu2006}, OU-MVLP\cite{Takemura2018}, GREW\cite{zhu2021gait} and Gait3D \cite{zheng2022gait3d}, GaitLU-1M is superior in the following aspects.

a. \textit{Scale.} GaitLU-1M makes one of the first attempts toward large-scale self-supervised learning for gait recognition, 
and its size is completely comparable to other visual tasks using contrastive methods, \textit{e.g.}, ImageNet\cite{deng2009imagenet} for image classification.

b. \textit{Diversity.} Since the chosen 1,379 videos are filmed around the world, varying from the city, place to time, 
it is almost impossible for the same person appears in two different videos in GaitLU-1M.
Hence we can regard the 1M walking sequences as 1M identities, 
implying that GaitLU-1M possesses millions of individual walking patterns.
Similarly, the 1,379 videos can be considered captured from 1,379 different cameras, varying from device to photographer.
These diversities offer excellent potential for our baseline model GaitSSB to learn the general gait representation.

c. \textit{Massive Variations.} Though the latest Gait3D\cite{zheng2022gait3d} and GREW\cite{zhu2021gait} are collected in complex environments (supermarkets and other particular areas),  
we consider that their diversities of variations are still less than GaitLU-1M since the latter is captured in worldwide practical scenarios.

\textcolor{black}{
Besides, two limitations of GaitLU-1M are worth mentioning: 
a. Since all the gait sequences are anonymous with the random index, GaitLU-1M has no weak annotations like filming cities. Also, the representations of silhouette and skeleton lack visual clues, making it hard for further attribute annotations. 
b. Most videos are captured by hand-held phones/cameras, which may cause the view distribution gap between GaitLU-1M and typical CCTVs, especially for tilt and pitch angles. 
In Sec.~\ref{sec3.2.2}, the perspective transformation strategy can alleviate this issue to some extent.
}

\subsection{GaitSSB}
Two gait-specific issues are making GaitSSB distinct from other contrastive frameworks,
\textit{i.e.}, the input modality and task granularity.
The first one requires a new data augmentation strategy concentrating on gait silhouette sequences, 
and the latter reminds us to learn the subtle yet discriminative gait's unique features.
In this subsection, 
we first describe the high-level design of GaitSSB.
Then, we present more details involving the silhouette augmentation operator and gait feature encoder.
Moreover, we also introduce the prevailing predictor and stop-gradient trick.
In the end, we compare GaitSSB with other contrastive frameworks from the perspectives of fine-grained and view-invariant gait representation learning.

\begin{figure*}[htbp]
\centering
\includegraphics[height=4.5cm]{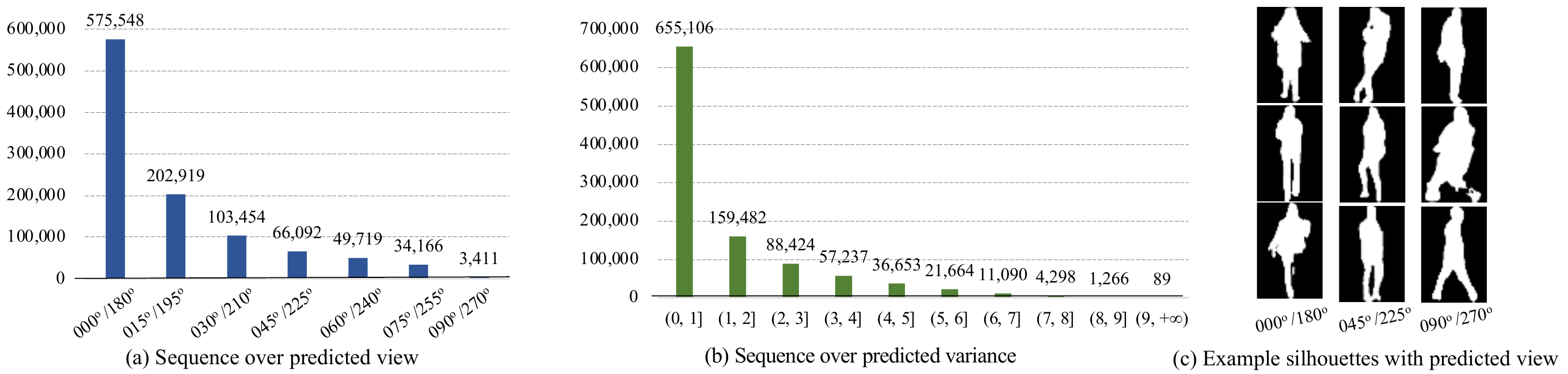}
\caption{(a) shows the statistics of sequences number over the predicted views. (b) shows the statistics of sequences number over the predicted variances. (c) shows example silhouettes with the predicted view.}
\label{fig5}
\vspace{-1.0em}
\end{figure*}

\begin{figure}[htbp]
\centering
\includegraphics[height=4.5cm]{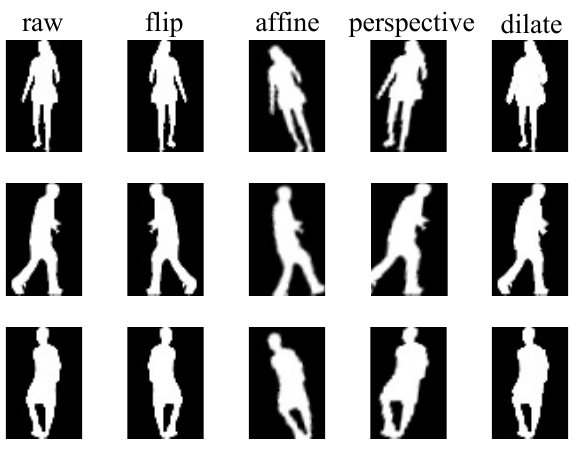}
\caption{Example silhouettes with spatial augmentation. From left to right: raw image, horizontal flipping, affine transformation including rotation, perspective transformation and dilation.}
\label{fig4}
\vspace{-1.0em}
\end{figure}

\subsubsection{Description of GaitSSB}
\label{sec3.2.1}
Our goal is to learn the general gait representation without labeling samples.
Inspired by successful contrastive learning, 
the core idea is to reduce the distance between two different augmented views of the same gait sequence (positive pairs) and simultaneously repulse the negative pairs, where the assumption is that data augmentation can simulate the practical variations to some extent without damaging identity information.

As shown in Fig.~\ref{fig3}, we sample two of the non-overlap clips ($c_i$ and $c_{i^+}$) in the sequence $S_i$ as two augmented views of identical walking patterns, where $i\in\left \{1, ..., n  \right \}$.
Hence, there are $2n$ gait clips in a training batch with $n$ positive pairs.
After being randomly augmented by silhouette augmentation operator (\textbf{SAO}), 
these clips will be fed into the gait feature encoder $\textrm{F}_\theta (\cdot)$ to extract the feature vectors, that is $k_i=\textrm{F}_\theta (c_i)$ and $k_{i^+}=\textrm{F}_\theta (c_{i^+})$.
\textcolor{black}{
Notably, the encoder $\textrm{F}_\theta (\cdot)$ is a part-based model, meaning that the output $k_i$ is a group of parallel feature vectors responding to different parts of the human body. 
Since these part vectors will be processed independently, the following formulation loosely treats $k_i$ as a single feature vector for brevity. 
}

Then, the predictor head, denoted as $\textrm{H}_\theta (\cdot)$, transforms the output of one view
and matches it to the another view, denoted as $q_i=\textup{H}_\theta(k_i)\rightarrow k_{i^+}$.
Consider $(q_i, k_{i^+})$ as the positive pair and ($q_i, k_{j^+}$) with $j\neq i$ as negative pairs, 
we formulate the InfoNCE loss\cite{gutmann2010noise}:
\begin{equation}
\label{equ1}
L(q_i, k_{i^+})=-\textup{log}\frac{\textup{exp}(\textup{sim}(q_i, k_{i^+})/\tau )}{\sum_{j=1}^{n}\textup{exp}(\textup{sim}(q_i, k_{j^+})/\tau)}
\end{equation}
where $\textup{sim}(q_i, k_{j^+}) = q_i^{\mathsf{T}}k_{j^+}/(\left \| q_i \right \|_2\left \| k_{j^+} \right \|_2)$ is the cosine similarity, 
and $\tau>0$ denotes the temperature hyper-parameter set to 16.

Following \cite{simsiam}, we use a symmetrized loss as:
\begin{equation}
\label{equ2}
\pounds=\frac{1}{2}L(q_i, k_{i^+}) + \frac{1}{2}L(q_{i^+}, k_i)
\end{equation}
where $q_{i^+}=\textup{H}_\theta(k_{i^+})$. 
This loss is defined for each gait clip, and the total loss is averaged over all clips in the training batch.
\textcolor{black}{
Moreover, the stop-gradient operation alternatively treats $k_i$ and $k_{i^+}$ as constant to prevent collapsing solutions\cite{simsiam}. 
Fig.~\ref{fig3} only shows the latter case for brevity. 
}

\subsubsection{Silhouette Augmentation Operator}
\label{sec3.2.2}
Data augmentation plays a vital role in contrastive framework because it can partially simulate the practical variations without damaging identity information.
\begin{figure}[htbp]
\centering
\includegraphics[height=1.8cm]{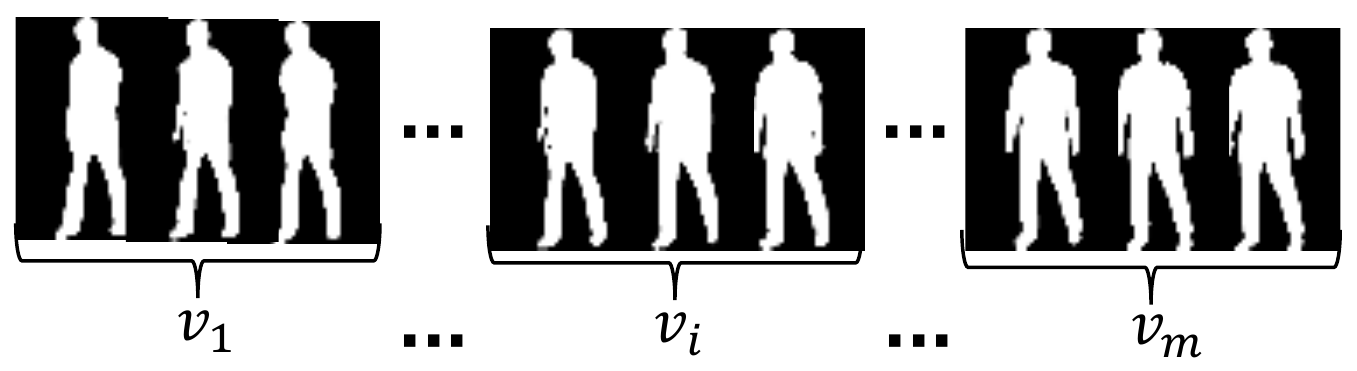}
\caption{A gait sequence may involve several clips captured from visually different camera angles.}
\label{fig-sampling-augmentation}
\vspace{-1.0em}
\end{figure}
However, 
since our input modality is the sequence of silhouettes,
the well-explored data augmentation methods mostly centralized in RGB images no longer apply to our gait contrastive framework. 
In this paper, 
we propose a novel silhouette augmentation operator (\textbf{SAO}, $\pi(\cdot)$) with making following three-fold effects: spatial, intra-sequence and sampling augmentations.

\textit{Spatial Augmentation.}
Performed on each frame, 
spatial augmentation simulates clothing/carrying changing factors.
Here we list several common operations centralizing in RGB image augmentation and discuss whether they still work for silhouette data.

\begin{itemize}
\item \textit{Inapplicable: Color distort.} The silhouettes are binary masks with no RGB information.
\item \textit{Unnecessary: Crop and resize, and Cutout.} Raw videos are detected and segmented frame by frame, suggesting GaitLU-1M dominates enormous natural occlusion and detection box drift.
\item \textit{Available: Flip, Rotate, Affine and Perspective.} These transformations with appropriate settings are still indispensable for silhouette augmentation.
\end{itemize}

Based on the above analysis, we adopt several augmentation methods, 
including horizontal flipping, slight rotation, affine and perspective transformations.
In addition, we introduce an extra dilation operator trying to simulate the most challenging clothing-changing factors.
Its core idea is to dilate a random consecutive region of the human body by randomly choosing structuring elements and kernel size, 
making the appearance look like wearing an additional coat.
Fig.~\ref{fig4} shows the instances illustrating the spatial augmentation used for GaitSSB.

\textit{Intra-sequence Augmentation.}
As mentioned in Sec.~\ref{sec3.2.1}, 
we consider two non-overlap silhouette clips inside the same walking sequence as two augmented views of identical walking patterns.
This strategy enables the model to learn gait feature consistency hidden in the temporal dimension\cite{qian2021spatiotemporal, li2021motion, dave2022tclr, feichtenhofer2021large}.
Ablation studies in Sec.~\ref{sec4} and discussions in Sec.~\ref{sec5} will declare the significance of temporal consistencies for the self-supervised gait pre-training.

\textit{Sampling Augmentation.}
We observe that frontal-view gait sequences possess almost no viewpoint transitions while others own relatively remarkable viewpoint changes inside the sequence.
Hence, to enhance the learning of view-invariant gait features, we propose a sampling strategy to choose these gait sequences possessing view-changing factors with a higher probability.

As shown in Fig.~\ref{fig-sampling-augmentation},
consider all the clips extracted from a walking sequence, 
we let the average of their camera angle, $\bar{v}=\frac{\sum_{i}^{m}v_i}{m}$, present the view of the entire sequence, 
and use the variance to quantitatively measure the degree of view transitions inside this sequence, denoted as $\sigma ^2=\frac{\sum_{i}^{m}(v_i-\bar{v})^2}{m-1}$.
The larger $\sigma ^2$ indicates the more salient view changes and vice versa. 
Specifically, 
a small multiple layers perceptron composed of a stack of four fully-connected layers (equipped with the batch normalization and ReLU activation layers),
and an extra classification layer is trained on OU-MVLP to serve as the view classifier.
We use the softmax loss with label smoothing to drive the training process.
It is worth noting that we tag some view pairs close to being horizontally symmetric with the same label to reduce the difficulty of convergence caused by the highly similar appearances.
Specifically, 
we tag both the original $000^\circ$ and $180^\circ$ as $\#0$, $015^\circ$ and $195^\circ$ as $\#1$, and so on
\footnote{The experimental results present over $95\%$ accuracy in predicting the camera angle of sequences with an error of $\pm 15^\circ$ on OU-MVLP.}.

Consequently, 
we can get more statistical analysis about GaitLU-1M.
As shown in Fig.~\ref{fig5} (a), 
we observe that about one-half sequences in GaitLU-1M are captured in frontal-view cases, \textit{i.e.}, those sequences under $000^\circ, 015^\circ, 180^\circ$ and $195^\circ$ camera angles.
On the other hand, 
there are over one-half of sequences in GaitLU-1M that own almost no view transitions, 
namely, those sequences with their $\sigma ^2 \in \left ( 0,1  \right ]$ as shown in Fig.~\ref{fig5} (b).
We name them by dumb gait sequences for convenience since they tell us no view-changing information.
Further statistics show that $85.6\%$ dumb sequences come from frontal-view cases.
Based on the above empirical analysis, 
we take a simple mining strategy at the pre-training phase: \textcolor{black}{sample non-dumb sequences with a large probability} to enhance view-invariant gait feature learning.

\begin{table}[t]
\centering
\caption{Architecture of gait feature encoder $\textup{F}_\theta(\cdot)$. The strings following each layer present the size and channel of the convolution kernels or the dimension of the separate fully-connected layers.}
\label{tab2}
\begin{tabular}{c|c|c}
\toprule
\textbf{Component}                        & \textbf{Layer}      & \textbf{Architecture}        \\ \midrule 
\multirow{10}{*}{Backbone}        & ConvLayer & $3\times3$, 64, stride=1   \\
\multirow{3}{*}                                 & ResBlock 1    & $\begin{bmatrix}
3\times3, 64\\ 
3\times3, 64
\end{bmatrix} \times 1, \textup{stride}=2$ \\
\multirow{3}{*}                                  & ResBlock 2    & $\begin{bmatrix}
3\times3, 128\\ 
3\times3, 128
\end{bmatrix} \times 1, \textup{stride}=2$ \\
\multirow{3}{*}                                  & ResBlock 3    & $\begin{bmatrix}
3\times3, 256\\ 
3\times3, 256
\end{bmatrix} \times 1, \textup{stride}=1$ \\
\multirow{3}{*}                                  & ResBlock 4    & $\begin{bmatrix}
3\times3, 512\\ 
3\times3, 512
\end{bmatrix} \times 1, \textup{stride}=1$ \\ \hline
TP                               & \multicolumn{2}{c}{Temporal Pooling}    \\ \midrule 
HP                               & \multicolumn{2}{c}{Horizontal Pooling} \\ \midrule 
\multirow{2}{*}{Head} & Separate FC 0       &  $512\times512$                   \\
                                 & Separate FC 1       &  $512\times512$                    \\ \bottomrule 
\end{tabular}
\vspace{-1.0em}
\end{table}
\subsubsection{Gait Feature Encoder}
Gait feature encoder $\textrm{F}_\theta (\cdot)$ aims to transform the input sequence into gait feature vectors.
In practice, it can be any existing methods, such as GaitSet\cite{Chao2019}, GaitPart\cite{fan2020gaitpart} and GaitGL\cite{gaitgl}.
However, we fail to introduce them and meet the under-fitting cases.
One possibility is that their networks are too shallow to fit millions of gait sequences. 
Therefore, we build a deep network for gait feature extraction, although that is not the main target of this paper.

Based on the review of recent typical works\cite{Chao2019, fan2020gaitpart, gaitgl}, 
we summarize that, 
to extract a good representation of gait sequence, 
four essential components are necessary for the encoder, 
\textit{i.e.}, convolution backbone, temporal pooling (TP) module, horizontal pooling (HP) module and projection head.
We construct a deep ResNet-like\cite{he2016deep} network as the backbone. 
Besides that, to make it concise enough, we decide not to utilize any other well-designed modules proposed in existing approaches\cite{Chao2019, fan2020gaitpart, gaitgl}, 
such as
Multilayer Global Pipeline in GaitSet\cite{Chao2019}, 
Focal Convolution layer in GaitPart\cite{fan2020gaitpart} and Global/Local Convolutional layer in GaitGL\cite{gaitgl}.

More specifically, as illustrated in Table~\ref{tab2}, 
the backbone is composed of one initial convolution layer followed by a stack of four basic residual blocks, 
devoting to transforming each silhouette of the input gait clip into a frame-level feature map.
And then, 
the temporal pooling (TP) module compresses the clip of feature maps along the temporal dimension by maximum function, 
thus outputting a feature map dominating clip-level understandings.
Next, 
the horizontal pooling (HP) module slices the obtained clip-level feature map horizontally into pre-defined parts, 
\textit{i.e.}, 16 partial feature map uniform in size.
Each part-level feature map will be down-sampled into a part-informed vector by global average and max pooling operations.
Finally, 
those extracted vectors will be mapped into the metric space by a projection head, 
who consists of two separate fully-connected layers and treats each part vector independently.
Notably, 
all the layers in Table~\ref{tab2} are equipped with the batch normalization and ReLU activation layers except the last output layer, 
and Table~\ref{tab2} skips them for conciseness.

Totally, 
these designs enable our model to hold the fitting of million gait sequences, 
making it structurally concise yet experimentally strong to serve as the baseline model for learning the general gait representation from massive unlabelled walking videos.

\subsubsection{Predictor and Stop-gradient}
The predictor head $\textrm{H}_\theta (\cdot)$ transforms the output of one view and matches it to another view, that is $q_i=\textup{H}_\theta(k_i)\rightarrow k_{i^+}$.
Its structure consists of two separate fully-connected layers, with just the first one equipped with the batch normalization and ReLU activation layers, 
which is similar to the mentioned projection head in the gait feature encoder.
Besides, 
a stop-gradient operation in Fig.~\ref{fig3} is introduced in~\cite{simsiam, grill2020bootstrap}, 
and aims at preventing the model's outputs from collapsing to constant vectors.

\begin{table}[htbp]
\caption{The amount of the identities (\#ID) and sequences (\#Seq) covered by the employed datasets. 
}
\label{tab3_1}
\begin{tabular}{cccccc}
\toprule
\multirow{2}{*}{Dataset} & \multicolumn{2}{c}{Train Set} & \multicolumn{2}{c}{Test Set} & \multirow{2}{*}{Condition} \\
                         & \#Id            & \#Seq           & \#Id           & \#Seq           &                         \\ \midrule 
CASIA-B                  & 74            & 8,140         & 50           & 5,500         & NM, BG, CL              \\
OU-MVLP                   & 5,153         & 144,284       & 5,154        & 144,312       & NM                      \\
GREW                     & 20,000        & 102,887       & 6,000        & 24,000        & Diverse                   \\
Gait3D                   & 3,000         & 18,940        & 1,000        & 6,369         & Diverse                   \\ \bottomrule 
\end{tabular}
\vspace{-1.0em}
\end{table}

Empirically, 
some works\cite{simsiam, grill2020bootstrap} formulate the learning process of predictor $\textrm{H}_\theta (\cdot)$ as an optimization problem that is being solved underlying.
By definition, the optimal solution to $\textrm{H}_\theta (\cdot)$ should satisfy: 
$q_i=\textrm{H}_\theta (k_i)=\mathbb{E_\pi}(k_{i^+})$, 
indicating that the predictor head $\textrm{H}_\theta (\cdot)$ is an approximation of the expectation $\mathbb{E_\pi}(\cdot)$ over the data augmentation involved by $\pi(\cdot)$.
This hypothesis and other similar interpretations \cite{chen2020improved, grill2020bootstrap, simsiam, caron2020unsupervised} can declare the necessity of the predictor $\textrm{H}_\theta (\cdot)$, that is to overcome the variations caused by $\pi(\cdot)$.

\subsubsection{Comparison with Other Contrastive Frameworks}
By ignoring details in each particular component, 
here we compare our GaitSSB with other contrastive frameworks in design concepts.

Interestingly, 
if we remove the negative samples, 
in other words, 
replace InfoNCE\cite{oord2018representation} with cosine similarity loss\cite{simsiam, grill2020bootstrap}, 
GaitSSB would be similar to the SimSiam\cite{simsiam} equipped with some modifications for fitting the gait-specific issues.
However, 
we argue that the task granularity makes it different a lot.
More specifically, 
gait recognition is a fine-grained task whose main challenge is that intra-class distances are often larger than inter-class, 
especially for cross-view cases.
Once we discard the explicit signal of increasing the distance between representations of different gait sequences, 
those sequences owning similar appearance because of the similar camera angle would be automatically clustered together since the overall visual similarity often dominates in contrastive frameworks\cite{Wu_2018_CVPR, cole2021does}.
This may can cause a massive obstacle in identifying individuals because the extracted features take the view characteristics as centers.
Hence, 
we think that the InfoNCE loss, which not only reduce the distance between representations of positive pairs but also enlarge that of negative pairs, 
is crucial for the gait contrastive framework.
The ablation study in Sec.~\ref{sec4.5.2} and the hypothesis in Sec.~\ref{sec5} stand for this point experimentally and theoretically.

In addition, compared with those contrastive learning frameworks using InfoNCE loss, \textit{e.g.},  MoCo\cite{he2020momentum} and SwAV\cite{caron2020unsupervised}, 
we take a more concise siamese network\cite{chopra2005learning} as the encoder, 
instead of the relatively complex momentum or online clustering encoder.
The primary motivation comes from the philosophy that the simple one is enough.
In a nutshell, 
after absorbing the advantages of existing works, 
GaitSSB is conceptually simple and empirically powerful enough to be a self-supervised benchmark for gait recognition.

\section{Experiments}
\label{sec4}
This section first introduces the datasets in use, including our unlabelled GaitLU-1M for pre-training and four public supervised datasets for transfer learning.
Then, the implementation details about self-supervised pre-training, supervised transfer learning, and training from scratch will be reported.
In the end, 
we conduct systematical comparisons and comprehensive ablation studies, 
adequately exhibiting the superiority of GaitSSB and its components.

\subsection{Datasets}
GaitLU-1M collected by this paper is an unlabelled dataset used for pre-training.
As for transfer learning, 
we employ four public supervised gait datasets, 
involving the most widely-used CASIA-B\cite{Yu2006}, 
the largest indoor OU-MVLP\cite{Takemura2018}, 
the largest dataset in the wild GREW\cite{zhu2021gait}, 
and the most up-to-date popular Gait3D\cite{zheng2022gait3d}.
Here we describe implementation details that strictly follow the official protocols.

\textit{CASIA-B}\cite{Yu2006} is composed of 124 subjects, and each subject contains 3 walking conditions, including the normal (NM\#1-6), bag-carrying (BG\#1-2) and clothing-change (CL\#1-2) cases.
Each walking video is taken from 11 views uniformly distributed in [$0^{\circ}, 180^{\circ}$].
Hence, there are $11 \times (6 + 2 + 2) = 110$ sequences per subject.
In our experiments, the first 74 subjects are grouped into the training set, while the remaining 50 are grouped for evaluation.
At the test stage, for each subject, the first 4 sequences under the NM walking (NM\#1-4) are grouped as the gallery set, and the remaining 6 are divided into 3 probe subsets according to walking conditions, \textit{i.e.}, NM, BG and CL.

\textit{OU-MVLP}\cite{Takemura2018} is one of the largest indoor gait datasets. 
This dataset comprises 10,307 subjects, and each subject only contains one walking condition, namely normal walking (NM\#0-1).
Each walking video is taken from 14 views uniformly distributed inside [$0^{\circ}, 90^{\circ}$] and [$180^{\circ}, 270^{\circ}$].
Therefore, there are $ 2 \times 14 = 28$ sequences per subject ideally.
Following the official partition strategy, 
we take 5,153 subjects as the training set and 5,154 subjects as the test set.
At the test stage, each subject's first sequence is regarded as the probe, and another is used for the gallery.

\begin{table*}[htbp]
\centering
\caption{Rank-1 accuracy on four authoritative gait datasets: unsupervised performance of GaitSSB \textit{v.s.} supervised performance of some early methods.
}
\begin{tabular}{c|c|c|ccccccc}
\toprule 
\multirow{3}{*}{Mode}                                      & \multirow{3}{*}{Method} & \multirow{3}{*}{Srouce} & \multicolumn{7}{c}{Dataset}                                                                                                                                                                              \\ \cline{4-10} 
&                         &                         & \multicolumn{3}{c|}{CASIA-B}      & \multicolumn{1}{c}{\multirow{2}{*}{OU-MVLP}} & \multicolumn{1}{c|}{\multirow{2}{*}{\textcolor{black}{OU-MVLP$^\dag$}}} & \multicolumn{1}{c|}{\multirow{2}{*}{GREW}} & \multirow{2}{*}{Gait3D} \\
                                                           &                         &                         & NM                        & BG                        & \multicolumn{1}{c|}{CL}   & & \multicolumn{1}{c|}{}                         & \multicolumn{1}{c|}{}                      &                         \\ \midrule 
\multirow{4}{*}{Supervised}                                & PoseGait~\cite{liao2017pose}                & PR2020                  & 63.8                      & 42.5                      & \multicolumn{1}{c|}{32.0} & \multicolumn{1}{c}{-} & \multicolumn{1}{c|}{-}                        & \multicolumn{1}{c|}{0.2}                   & 0.2                     \\
                                                           & GaitGraph~\cite{teepe2021gaitgraph}               & ICIP2021                & 87.7                      & 74.8                      & \multicolumn{1}{c|}{\textbf{66.3}} & \multicolumn{1}{c}{-} &\multicolumn{1}{c|}{-}                        & \multicolumn{1}{c|}{1.3}                   & 6.2                     \\
                                                           & CNN-LB~\cite{wu2016comprehensive}                  & PAMI2015                & \textbf{94.1}                      & 72.4                      & \multicolumn{1}{c|}{54.0} &\multicolumn{1}{c}{-} & \multicolumn{1}{c|}{-}  & \multicolumn{1}{c|}{13.6}                  & -                       \\
                                                           & GEINet~\cite{Shiraga2016}                  & ICB2016                 & \multicolumn{3}{c|}{-}                                 &-                           & \multicolumn{1}{c|}{\textbf{35.8}}                     & \multicolumn{1}{c|}{6.8}                   & 7.0                     \\ \midrule 
\begin{tabular}[c]{@{}c@{}}Self-\\ Supervised\end{tabular} & GaitSSB                 & ours                    & \multicolumn{1}{c}{83.3} & \multicolumn{1}{c}{\textbf{75.6}} & \multicolumn{1}{c|}{28.7} & \textbf{37.2} & \multicolumn{1}{c|}{35.0}                     & \multicolumn{1}{c|}{\textbf{16.6}}                  & \textbf{24.7}                    \\ \bottomrule 
\end{tabular}
\label{tab4_1}
\vspace{-1.0em}
\end{table*}

\begin{table}[htbp]
\caption{Common training settings on various datasets at different stages.
The batch size of $(q, k)$ indicates $q$ subjects with $k$ sequences per subject.}
\label{tab3}
\begin{tabular}{ccccc}
\toprule
\multirow{2}{*}{Stage}                                                              & \multirow{2}{*}{DataSet} & \multirow{2}{*}{Batch Size} & \multirow{2}{*}{Milestones} & \multirow{2}{*}{Steps} \\
                                                                                    &                          &                             &                             &                        \\ \midrule 
Pre-training                                                                        & GaitLU-1M                & 512                         & (80K, 120K)                 & 150K                   \\ \hline
\multirow{4}{*}{\begin{tabular}[c]{@{}c@{}}Transfer\\ Learning\end{tabular}}        & CASIA-B                  & (8, 16)                     & (10K)                       & 12K                    \\
                                                                                    & OU-MVLP                   & (32, 16)                    & (50K, 60K, 70K)             & 80K                    \\
                                                                                    & GREW                     & (128, 4)                    & (50K, 60K, 70K)             & 80K                    \\
                                                                                    & Gait3D                   & (64, 4)                     & (6K, 8K, 10k)               & 12K                    \\ \midrule 
\multirow{4}{*}{\begin{tabular}[c]{@{}c@{}}Training\\ from \\ Scratch\end{tabular}} & CASIA-B                  & (8, 16)                     & (10K, 20K, 30K)             & 40K                    \\
                                                                                    & OU-MVLP                   & (32, 16)                    & (60K, 80K, 100K)            & 120K                   \\
                                                                                    & GREW                     & (128, 4)                    & (60K, 80K, 100K)            & 120K                   \\
                                                                                    & Gait3D                   & (64, 4)                     & (20K, 40K, 50K)             & 60K                    \\ \bottomrule 
\end{tabular}
\vspace{-1.0em}
\end{table}

\textit{GREW}\cite{zhu2021gait} is the largest gait dataset in the wild up to date to our best knowledge. 
Its raw videos are collected from 882 cameras in a large public area, containing nearly 3,500 hours of 1,080$\times$1,920 streams.
Except for the identities, some other human attributes have been annotated as well, 
\textit{e.g.}, the gender, 14 age groups, 5 carrying conditions and 6 dressing styles.
Therefore, GREW is believed to include adequate and diverse practical variations.
In addition, this dataset is divided into 4 parts: 
a train set with 20,000 identities and 102,887 sequences, a validation set with 345 identities and 1,784 sequences, a test set with 6,000 identities and 24,000 sequences, 
a distractor set with 233,857 anonymous sequences.
At the testing stage, 
each subject in the test set has 4 sequences, 2 for the probe and another 2 for the gallery.
We use no valuation and distractor set for convenience.

\textit{Gait3D}\cite{zheng2022gait3d} was collected in a large supermarket and contains 1,090 hours of videos with 1,920$\times$1,080 resolution and 25 FPS.
It consists of 4,000 subjects with over 25,000 walking sequences, 
in which 3,000/1,000 IDs are grouped into train/test subsets.
For the test set, 
only one sequence from each ID will be selected to build the probe set with 1,000 sequences, 
while the rest becomes the gallery set with 5,369 sequences.
For clarity, we list several key indexes for employing the above datasets in Table~\ref{tab3_1}.

\subsection{Implementation Details}
Unless otherwise stated, 
\textrm{a}) The input silhouettes are aligned by the size-normalization method used in \cite{Takemura2018} and resized to $64\times44$.
\textrm{b}) SGD optimizer with an initial learning rate of 0.1 and momentum of 0.9 is employed for training.
\textrm{c}) The multiple-step scheduler for dropping the learning rate with milestones shown in Table~\ref{tab3} is adopted.
\textrm{d}) The triplet loss\cite{Hermans2017} with a margin of 0.3 is employed at supervised learning stages.
\textrm{e}) The batch size and total iterations are listed in Table~\ref{tab3}.
\textrm{f}) The length of the input sequence is set to 16 frames in the pre-training phase and 30 frames for the other phases.
\textrm{g}) Several state-of-the-art methods are reproduced, involving the GaitSet\cite{Chao2019}, GaitPart\cite{fan2020gaitpart}, GaitGL\cite{gaitgl} and CSTL\cite{huang2021context} (with $128\times88$ input), 
because their performance on GREW\cite{zhu2021gait} and Gait3D\cite{huang20213d} are missing. 
\textrm{h}) Our code is based on the popular OpenGait~\cite{fan2022opengait} codebase available at \url{https://github.com/ShiqiYu/OpenGait}.

\subsubsection{Pre-training}
The particular settings for pre-training mainly center on the hyper-parameters of silhouette augmentation operator $\pi(\cdot)$.

For spatial augmentation, 
we randomly perform the horizontal flipping, affine, perspective and dilation transformations with a probability of 0.5.
The affine transformation consists of a rotation with a random angle from $[-10^\circ, +10^\circ]$, and a shear with its level ranging on $[-5\times10^{-3}, 5\times10^{-3}]$.
The transform axes of perspective are generated by randomly skewing the source axes within 10 pixels.
Besides, dilation operations are conducted on a random contiguous region in the silhouette with a random kernel shape (rectangle, cross, or ellipse) and size (3 or 5). 

For sampling augmentation, 
we sample the dumb sequences
with a small probability of 0.1, 
while the rest are for others to enhance the learning of view-invariant features. 

\subsubsection{Transfer Learning}
For transfer learning, 
we import an additional separate fully-connected layer, 
whose parameters are $l_2$-normalized over channels dimension, to map the feature vectors ($l_2$-normalized) output by gait feature encoder $\text{F}_\theta(\cdot)$ into the metric space.
Several common fine-tuning tricks are employed to avoid catastrophic forgetting, 
\textit{i.e.}, 
using the small initial learning rate, fixing the batch normalization layers, and removing the weight decay.
Specifically, 
the initial learning rate of backbone, projection and additional head is set to $1.0\times10^{-3}$, $1.0\times10^{-2}$ and $1.0\times10^{-1}$, respectively.

\subsubsection{Training from Scratch}
To intuitively show the effectiveness of pre-training, 
we take GaitSSB directly trained with random initialization into comparison scope.
As for the training settings, 
there are only two slight differences between this case with the transfer learning, 
\textit{i.e.}, starting with the learning rate of 0.1 and training with a new scheduler shown in Table~\ref{tab3}.

We take the widely-used average rank-1 accuracy as the metric.
Moreover, 
\textrm{a}) For CASIA-B, there are three metrics responding to the performance on its three probe subsets, \textit{i.e.}, NM, BG and CL.
\textrm{b}) For CASIA-B and OU-MVLP, the identical-view cases are excluded.
\textrm{c}) 
\textcolor{black}{
For OU-MVLP, those probe sequences owning no gallery are skipped by default during the evaluation.
Also, we make OU-MVLP$^\dag$ present the results without this skipping process. 
}

\subsection{Unsupervised Performance Comparison}
We compare the unsupervised performance of GaitSSB with the supervised performance of four early representative methods, 
including the model-based PoseGait\cite{liao2020model} and GaitGraph\cite{teepe2021gaitgraph}, and the GEI-based CNN-LB\cite{Wu2016} and GEINet\cite{shiraga2016geinet}, 
on four authoritative gait benchmarks, namely CASIA-B\cite{Yu2006}, OU-MVLP\cite{Takemura2018}, GREW\cite{zhu2021gait} and Gait3D\cite{zheng2022gait3d}.

\begin{table*}[htbp]
\centering
\caption{Rank-1 accuracy on four authoritative gait datasets: unsupervised performance of GaitSSB \textit{v.s.} cross-domain performance of recently state-of-the-art methods. The numbers in $\pmb{[}\quad\pmb{]}$ indicate the supervised results, and others present the cross-domain or unsupervised results. 
}
\centering
\includegraphics[height=7.0cm]{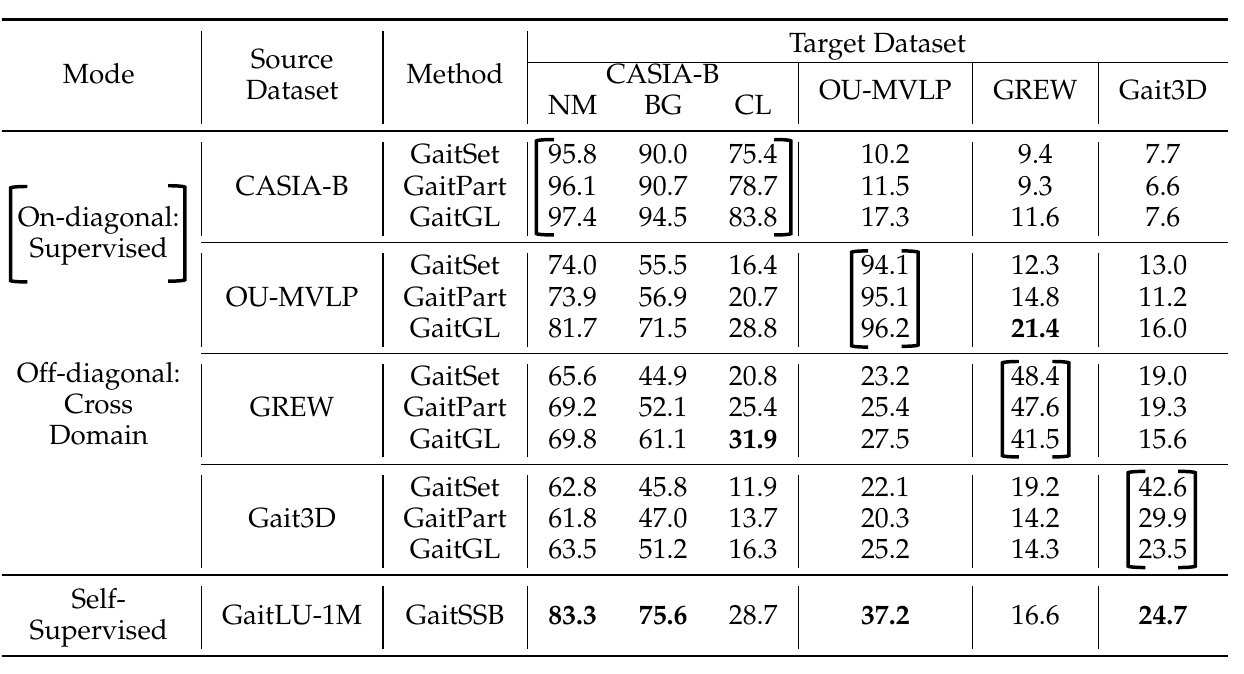}
\label{tab4_2}
\vspace{-1.0em}
\end{table*}

\subsubsection{Compared with Early Supervised Methods}
As shown in Table~\ref{tab4_1}, 
GaitSSB achieves nearly comparable or even better performance than the model-based (PoseGait and GaitGraph) and early GEI-based methods (CNN-LB and GEINet) in most cases, \textit{i.e.}, on the OU-MVLP, GREW and Gait3D, and in the NM, BG walking situations of CASIA-B.
However,
for the most challenging CL cases on CASIA-B, 
GaitSSB encounters an evident degradation, implying that the dress-changing factors stimulated by the silhouette augmentation operator $\pi(\cdot)$
are far from enough compared with those in the supervised datasets.

\subsubsection{Compared with Cross-Domain Testings}
To our knowledge, 
there are no reports about the unsupervised gait recognition performance evaluated on the aforementioned datasets.
For a comprehensive comparison,
we take the cross-domain testings into contrast scope due to its similarities with our evaluation protocol, 
\textit{i.e.}, train the model in supervised \textit{v.s.} self-supervised mode on the source dataset first and then test it on another target dataset.

As shown off-diagonal in Table~\ref{tab4_2}, 
our unsupervised performances remarkably outperform the cross-domain performance of existing supervised methods over most of the source-target dataset pairs, 
convincingly suggesting the robustness and generalization of the gait representation learned by our contrastive framework.
We think that the underlying reasons are two-fold: 
\textrm{a}) \textit{Diverse data.} The large-scale unlabelled gait dataset, GaitLU-1M, 
is far beyond the traditional gait datasets in terms of scale and variety, 
strongly supporting this impressive performance in the case of cross-domain.
\textrm{b}) \textit{Strong baseline.} GaitSSB can explore the general and discriminative representations for gait recognition without requiring labeling information.

\subsubsection{Cross-View Performance Analysis}
Furthermore, we are highly concerned about the unsupervised performance facing one of the most crucial and essential issues for practical gait recognition, 
namely the cross-view cases.
Hence, we draw the rank-1 accuracy heat map by taking the probe-gallery view pairs as coordinates based on the unsupervised results on CASIA-B and OU-MVLP.
Here we only consider the normal cases in CASIA-B to exclude the influence of other non-view factors.

\begin{figure}[htbp]
\centering
\includegraphics[height=5.0cm]{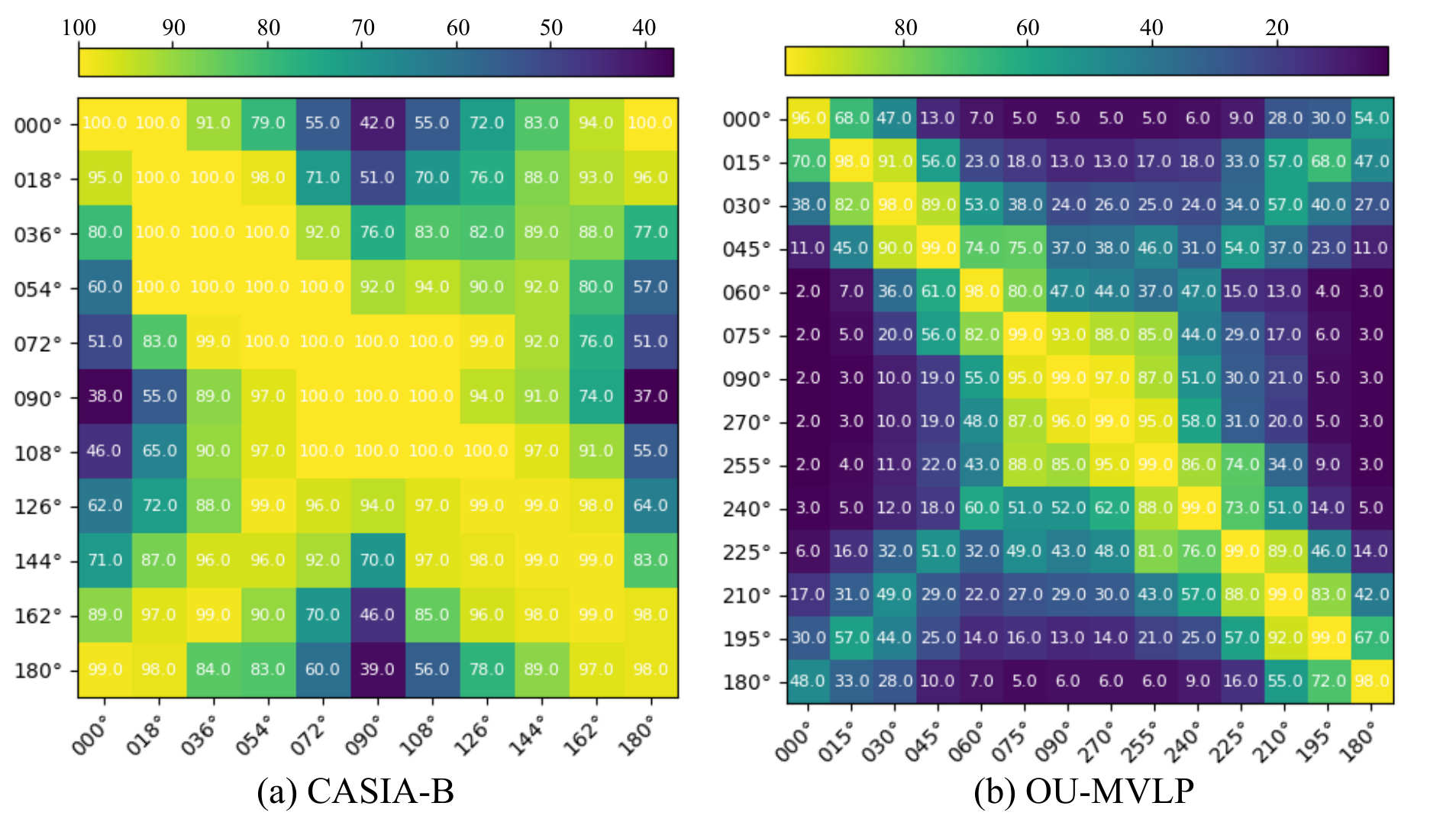}
\caption{GaitSSB's unsupervised rank-1 accuracy on (a) CASIA-B (NM walking) and (b) OU-MVLP over probe-gallery view pairs.}
\label{fig6}
\vspace{-1.0em}
\end{figure}

As shown in Fig.~\ref{fig6}, 
the results on the diagonal present the unsupervised performance in identical-view cases.
Encouragingly, 
GaitSSB makes a great achievement in these cases, 
\textit{i.e.} $99.5\%$ and $98.5\%$ rank-1 accuracy average over camera angles on CASIA-B and OU-MVLP, respectively, 
declaring that our unsupervised GaitSSB can nearly solve the identical-view gait recognition problem. 

\begin{table*}[htbp]
\caption{Rank-1 accuracy on five authoritative gait datasets: GaitSSB with transfer learning \textit{v.s.} some recently state-of-the-art methods.
}
\centering
\begin{tabular}{c|c|cccccccccc}
\toprule
\multirow{3}{*}{Method} & \multirow{3}{*}{Source} & \multicolumn{9}{c}{Dataset} \\ \cline{3-12} 
&  & \multicolumn{3}{c|}{CASIA-B}  & \multicolumn{3}{c|}{CASIA-B*} & \multicolumn{1}{c}{\multirow{2}{*}{OU-MVLP}} & \multicolumn{1}{c|}{\multirow{2}{*}{\textcolor{black}{OU-MVLP$^\dag$}}} & \multicolumn{1}{c|}{\multirow{2}{*}{GREW}} & \multirow{2}{*}{Gait3D} \\
&  & NM & BG & \multicolumn{1}{c|}{CL}   & NM   & BG   & \multicolumn{1}{c|}{CL}   &  & \multicolumn{1}{c|}{}  & \multicolumn{1}{c|}{}                      &                         \\ \midrule 
GaitSet~\cite{Chao2019}                 & AAAI 2019               & 95.8                           & 90.0                           & \multicolumn{1}{c|}{75.4} & 92.3 & 86.1 & \multicolumn{1}{c|}{73.4} & 94.1   & \multicolumn{1}{c|}{87.2}                 & \multicolumn{1}{c|}{48.4}                  & 42.6                    \\
GaitPart~\cite{fan2020gaitpart}                & CVPR2020                & 96.1                           & 90.7                           & \multicolumn{1}{c|}{78.7} & 93.1 & 86.0 & \multicolumn{1}{c|}{75.0} & 95.1   & \multicolumn{1}{c|}{88.6}                 & \multicolumn{1}{c|}{47.6}                  & 29.9                    \\
\textcolor{black}{KoopmanGait~\cite{Zhang_2021_CVPR}}                & \textcolor{black}{CVPR2021}                & \multicolumn{3}{c|}{-} & \multicolumn{3}{c|}{-} & -                     & \multicolumn{1}{c|}{74.7}                & \multicolumn{1}{c|}{-}  & -                    \\
GaitGL~\cite{lin2021gait}                  & ICCV2021                & 97.4                           & 94.5                           & \multicolumn{1}{c|}{83.8} & 94.1 & 90.0 & \multicolumn{1}{c|}{81.4} & 96.2 & \multicolumn{1}{c|}{89.9}                     & \multicolumn{1}{c|}{47.3}                  & 23.5                    \\
CSTL~\cite{huang2021context}                    & ICCV2021                & 98.0                           & 95.4                           & \multicolumn{1}{c|}{87.0} & 90.1 & 85.6 & \multicolumn{1}{c|}{79.2} & -  & \multicolumn{1}{c|}{90.2}                     & \multicolumn{1}{c|}{50.6}                  & -                       \\
3DLocal~\cite{3DLocal}                 & ICCV2021                & \textbf{98.3} & \textbf{95.5} & \multicolumn{1}{c|}{84.5} & \multicolumn{3}{c|}{-}                 &96.5 & \multicolumn{1}{c|}{90.9}                     & \multicolumn{1}{c|}{-}                     & -                       \\
\textcolor{black}{GaitMPL~\cite{dou2022gaitmpl}}                 & \textcolor{black}{TIP2022}                & 97.5 & 94.5 & \multicolumn{1}{c|}{\textbf{88.0}} & \multicolumn{3}{c|}{-}               & -   & \multicolumn{1}{c|}{90.6}                     & \multicolumn{1}{c|}{-}                     & -                       \\
SMPLGait~\cite{zheng2022gait3d}                & CVPR2022                & \multicolumn{3}{c|}{-}                                                                      & \multicolumn{3}{c|}{-}                  & -                        & \multicolumn{1}{c|}{-}                  & \multicolumn{1}{c|}{53.2}   & -                    \\ 
\textcolor{black}{LagrangeGait~\cite{chai2022lagrange}}                & \textcolor{black}{CVPR2022}                & 97.5                           & 94.6                           & \multicolumn{1}{c|}{85.1}                                                                      & \multicolumn{3}{c|}{-}               & -   & \multicolumn{1}{c|}{90.0}                        & \multicolumn{1}{c|}{-}                     & -                    \\ 
\textcolor{black}{GaitStrip~\cite{wang2022gaitstrip} }               & \textcolor{black}{ACCV2022}                & 97.6                           & 95.2                           & \multicolumn{1}{c|}{86.2}                                                                      & \multicolumn{3}{c|}{-}              & 97.0    & \multicolumn{1}{c|}{-}                        & \multicolumn{1}{c|}{-}                     & -                    \\ \midrule 
\multirow{2}{*}{GaitSSB}             & w/o pretraining   & 94.0                           & 87.6                           & \multicolumn{1}{c|}{68.0} & 96.6 & 93.1 & \multicolumn{1}{c|}{79.9} & 97.3 & \multicolumn{1}{c|}{91.0}                     & \multicolumn{1}{c|}{55.9}                  & 54.2                    \\
                 & w/ pretraining                        & 97.8                           & 93.9                           & \multicolumn{1}{c|}{77.4} & \textbf{96.8} & \textbf{94.5} & \multicolumn{1}{c|}{\textbf{81.8}} &\textbf{97.8} & \multicolumn{1}{c|}{\textbf{91.8}}                     & \multicolumn{1}{c|}{\textbf{61.7}}                  & \textbf{63.6}                    \\ \midrule 
\end{tabular}
\label{tab6}
\vspace{-1.0em}
\end{table*}

For the more challenging cross-view cases, 
whose results are displayed off-diagonal in Fig.~\ref{fig6}, 
\textcolor{black}{
there exist only two subcases for the unsupervised GaitSSB to reach an impressive over 50\% rank-1 accuracy on OU-MVLP. 
\textrm{a}) The view difference between probe and gallery is relatively small, \textit{e.g.}, 15$^\circ$ and 30$^\circ$, \textit{i.e.}, the results shown around the diagonal of Fig.~\ref{fig6} (b).
\textrm{b}) The appearance of the probe and gallery is close to being horizontally symmetric, \textit{i.e.} the results around the anti-diagonal of Fig.~\ref{fig6} (b).
}
But even so, GaitSSB exhibits competitive and promising results in some hard cases as well, 
\textit{e.g.}, 
$42.0\%$ rank-1 accuracy in the $000^\circ$-$090^\circ$ case for CASIA-B,
and $24.0\%$ rank-1 accuracy in the $030^\circ$-$090^\circ$ case for OU-MVLP.
In Sec.~\ref{sec5},
we will provide an instructive hypothesis about why and how our framework conquers such difficult view-changing factors.

Compared with the supervised results in the diagonal of Table~\ref{tab4_2}, 
there remains a significant gap in front GaitSSB's unsupervised performance compared with the supervised performance of existing state-of-the-art methods, 
especially for the dramatic dress/view-changing conditions. 
However, 
instead of exceeding the supervised methods on some particular datasets, 
the primary goal of GaitSSB is to learn the informative prior knowledge from the large-scale unlabelled walking videos for the more generalizable performance achieved in downstream tasks.

\subsection{Transfer Learning Performance}
After pre-training GaitSSB on GaitLU-1M, 
we fine-tune it on five target datasets, namely CASIA-B, CASIA-B*, OU-MVLP, GREW and Gait3D, to check its effectiveness and generalization for transfer learning.
Among these datasets, 
CASIA-B* is a re-segmented version~\cite{liang2022gaitedge}.
The extra import of CASIA-B* owes to the fact that the background subtraction algorithm that CASIA-B uses for generating silhouette data tends to produce much noise and is outdated for real-world applications nowadays.
\textcolor{black}{
In addition, to directly reveal the effectiveness of pre-training, 
we also evaluate the GaitSSB without pre-training, in other words, with being trained from scratch to exclude the backbone influence.
}

As shown in Table~\ref{tab6}, 
except on the CASIA-B, 
our GaitSSB reaches the new state-of-the-art performances on other four datasets, namely CASIA-B* (NM: $+2.7\%$, BG: $+4.5\%$, CL: $+0.4\%$), OU-MVLP ($+0.5\%$), GREW ($+9.7\%$) and Gait3D ($+10.4\%$).
It is worth noting that GaitSSB exceeds other methods by a larger margin on the datasets captured from real-world scenarios (GREW and Gait3D) instead of the controlled laboratory (CASIA-B* and OU-MVLP), 
illustrating two issues:
\textrm{a}) \textit{Generalization}. It is hard for other methods to transfer their superiority to practical cases.
\textrm{b}) \textit{Practicability}. GaitSSB is suitable for both indoor and outdoor cases.

For the confusing results on CASIA-B, 
we think that the salient cross-domain problem may cause it since the pedestrian segmentation algorithm that CASIA-B uses differs a lot from that of GaitLU-1M's, namely background subtraction \textit{v.s.} deep-learning based segmentation~\cite{liu2021paddleseg}.
For further analysis, 
we provide another two experiments to verify the effectiveness of our framework on CASIA-B: 
\textrm{a}) \textit{w/ v.s. w/o pre-training.} 
The experiment that initializes GaitSSB from the pre-trained model performs better than that from scratch.
\textrm{b}) \textit{CASIA-B* v.s. CASIA-B.} Once we re-segment the videos, \textit{i.e.}, experiments on CASIA-B*, the GaitSSB exceeds other methods simultaneously.

\begin{table}[t]
\caption{\textcolor{black}{Rank-1 accuracy under cross-domain settings.}}
\label{generalization}
\begin{tabular}{cc|cc}
\toprule
\multicolumn{1}{c|}{\multirow{2}{*}{Method}}  & \multirow{2}{*}{Source} & \multicolumn{2}{c}{Cross-domain: Source$\to$Target}                   \\ \cline{3-4} 
\multicolumn{1}{c|}{}                         &                         & GREW$\to$Gait3D     & Gait3D$\to$GREW   \\ \midrule 
\multicolumn{1}{c|}{GaitSet}                  & AAAI2019                & 19.0                          & 19.2                        \\
\multicolumn{1}{c|}{GaitPart}                 & CVPR2020                & 19.3                          & 14.2                        \\
\multicolumn{1}{c|}{GaitGL}                   & ICCV2021                & 15.6                          & 14.3                        \\ \midrule 
\multicolumn{1}{c|}{\multirow{2}{*}{GaitSSB}} & w/o pre-training        & 20.1                          & 20.6                        \\
\multicolumn{1}{c|}{}                         & w/ pre-training         & \textbf{27.2}                          & \textbf{27.6}                        \\ 
\bottomrule 
\end{tabular}
\vspace{-1.0em}
\end{table}

Notably, 
we also observe that excluding the CASIA-B and CASIA-B*, the performances of training GaitSSB from scratch, \textit{i.e.}, without pre-training, are distinctly beyond most of the current state-of-the-art methods on various benchmarks,
demonstrating that exploring basic network structures for gait recognition is far from enough.
The reason may be that the gait recognition community always relies too much on a gait dataset proposed before the boom of deep learning, 
namely, the CASIA-B, whose scale is pretty small and the pretreatment is out-of-date, from now view.

GaitSSB is a strong baseline model for gait recognition thanks to its simple structure and exceptional performance, even if we train it from scratch.
More importantly, the large-scale self-supervised pre-training makes our GaitSSB forward a step to bring the gait recognition an informed initialization, namely, the prior knowledge capacity to describe the gait patterns.

\textcolor{black}{
\subsubsection{Generalization Capacity Validation}
To directly validate the generalization capacity brought by the pre-training, we introduce Table~\ref{generalization} showing that, under the challenging cross-domain settings, 
the pre-training provides a noticeable performance gain for GaitSSB to outperform prior representative state-of-the-art methods by a margin. 
As we can see, the superior generalization capacity of GaitSSB mainly comes from the pre-training process instead of its architectural designs. 
}

\begin{figure}[t]
\centering
\includegraphics[height=3.0cm]{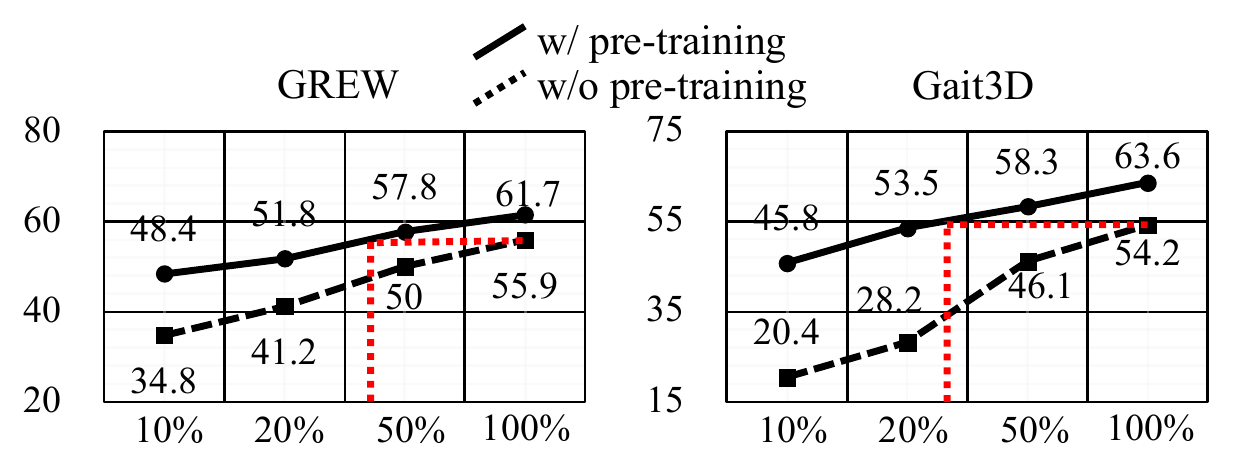}
\caption{\textcolor{black}{
Rank-1 accuracy of GaitSSB w/ \textit{v.s.} w/o pre-training when the fine-tuning (training) scale varies.}}
\label{finetune}
\vspace{-1.0em}
\end{figure} 

\begin{table*}[htbp]
\centering
\caption{Ablation studies: w/ and w/o the spatial, intra-sequence and sampling augmentation. w/ and w/o the negative samples.}
\begin{tabular}{c|cccc|ccc|c|c|c}
\toprule
\multirow{2}{*}{Index} & \multicolumn{4}{c|}{Control Variables}                                                                        & \multicolumn{3}{c|}{CASIA-B*}                                                                    & \multirow{2}{*}{OU-MVLP}        & \multirow{2}{*}{GREW}          & \multirow{2}{*}{Gait3D}        \\ \cline{2-5}
                       & Spatial Aug               & Intra-Sequence Aug        & Sampling Aug              & Negative Samples          & NM                             & BG                             & CL                             &                                &                                &                                \\ \midrule 
(\textbf{a})                    & $\checkmark$ & $\checkmark$ & $\checkmark$ & $\checkmark$ & \textbf{75.0} & \textbf{65.3} & \textbf{28.7} & \textbf{37.2} & 16.6                           & \textbf{24.7} \\ \midrule 
(\textbf{b})                    & $\times$                  & $\checkmark$              & $\checkmark$              & $\checkmark$              & 67.4                           & 56.7                           & 21.6                           & 26.7                           & 10.9                           & 18.5                           \\
(\textbf{c})                    & $\checkmark$              & $\times$                  & $\times$                  & $\checkmark$              & 6.9                            & 5.6                            & 3.4                            & 0.1                            & 0.1                            & 0.1                            \\
(\textbf{d})                    & $\checkmark$              & $\checkmark$              & $\times$                  & $\checkmark$              & 67.5                           & 58.6                           & 25.5                           & 33.5                           & \textbf{17.0} & 24.2                           \\
(\textbf{e})                    & $\checkmark$              & $\checkmark$              & $\checkmark$              & $\times$                  & 43.4                           & 32.1                           & 13.2                           & 14.6                           & 8.8                            & 15.5                           \\ \bottomrule 
\end{tabular}
\label{AblationStudy}
\vspace{-1.0em}
\end{table*}

\textcolor{black}{
\subsubsection{Annotation Cost Saving}
Fig.~\ref{finetune} shows the performance of GaitSSB w/ \textit{v.s.} w/o pre-training when the fine-tuning (training) scale varies. 
In practice, the first 10\%, 20\%, 50\% and 100\% subjects of the training set will be employed to fine-tune the pre-trained GaitSSB or directly train GaitSSB from scratch. 
Due to scale differences, we proportionally reduce the iterations used for dropping the learning rate and stopping training with keeping other settings shown in Table~\ref{tab3} unchanged. 
Consequently, Fig.~\ref{finetune} illustrates that: 
a) The smaller fine-tuning (training) scale, the more significant improvements brought by pre-training. 
b) According to the red dotted lines, compared with the full performance of GaitSSB without per-training, GaitSSB with pre-training reaches the competitive performance using only 50\% and 20\% training data of GREW and Gait3D, namely, the pre-training on the GaitLU dataset can save about 50\% and 80\% annotation costs of GREW and Gait3D.
}

\subsection{Ablation Study}
\label{sec4.5}
Unlike previous methods that only conduct the ablative studies on the fairly small-scale CASIA-B\cite{Chao2019, fan2020gaitpart, gaitgl, CSTL, 3DLocal}, 
here we perform the experiments on four datasets, 
including the CASIA-B*, OU-MVLP, GREW and Gait3D, 
to make our conclusions more solid.
Moreover, 
various controlling settings have been taken into consideration, involving w/ and w/o spatial, intra-sequence and sampling augmentations, 
w/ and w/o negative samples in the training batch, 
and different scales of pre-training datasets.
As shown in Table~\ref{AblationStudy}, 
we number the experiments for description convenience, where experiment~(\textbf{a}) denotes the complete GaitSSB. 

\subsubsection{Impact of Silhouette Augmentation}
\label{AblationStudy.1}
In this paper, 
the proposed silhouette augmentation operator $\pi(\cdot)$ consists of three parts, 
involving the spatial, intra-sequence and sampling augmentations.
Next, we will discuss their roles for GaitSSB, respectively. 

If we remove the spatial augmentation from GaitSSB, 
\textit{i.e.}, experiment (\textbf{a}) \textit{v.s.} (\textbf{b}) shown in Table~\ref{AblationStudy}, 
the performance would drop dramatically over all the datasets.
This phenomenon shows that the proposed spatial augmentation strategy, 
which is composed of the horizontal flipping, affine, perspective and dilation transformations, 
can simulate the spatial variations usually seen in silhouettes adequately and thus boost the pre-training performance remarkably.

The comparison between experiment (\textbf{a}) and (\textbf{c}) in Table~\ref{AblationStudy} indicates that the performance would collapse once we discard the intra-sequence and sampling augmentation.
In other words, 
GaitSSB cannot depict discriminative gait features by only taking the spatial augmentation.
By taking two non-overlap clips inside the same walking sequence as two different augmented views, 
the intra-sequence augmentation gives GaitSSB the ability against frame-level pretreatment errors, pose and appearance changes, 
and cross-view factors along the sequence dimension: 
\begin{itemize}
\item Since the natural walking videos are processed frame by frame, learning the intra-sequence consistencies enables the model to overcome the massive noises caused by pedestrian tracking and segmentation.
\item Two non-overlap clips inside the same walking sequence present identical gait patterns with two different views of appearance and pose.
Therefore, the intra-sequence consistencies imply the uniqueness of an individual's gait patterns to some extent, which is fundamental and crucial for gait recognition.
\item Walking is a dynamic course. The angle between the individual's walking route and the camera's orientation is not constant.
Hence, reducing the distance between representations of two non-overlap clips can benefit the learning of view-invariant features.
\end{itemize}

As for the sampling augmentation, 
its goal is to enhance the learning of view-invariant gait features via sampling the gait sequences possessing view-changing factors with a higher probability and vice versa.
By comparing experiment (\textbf{a}) with (\textbf{d}) in Table~\ref{AblationStudy}, 
we find this strategy benefits GaitSSB on various benchmarks observably, 
\textit{i.e.}, CASIA-B* (NM: $+7.5\%$, BG: $+6.7\%$, CL: $+3.2\%$), OU-MVLP ($+3.7\%$) and Gait3D ($+0.5\%$).
However, we also investigate that the sampling augmentation leads to a slight degradation in GREW.
\textcolor{black}{
Due to its largest scale and in-the-wild settings, identifying individuals on GREW is more difficult than other employed gait datasets.
This situation may make the improvement brought by sampling augmentation during the pre-training hard to be transferred to the downstream unsupervised evaluation on GREW. 
We consider that this strategy can boost pre-training performance in most cases, and meanwhile, a better sampling strategy is needed.
}

\subsubsection{Impact of Negative Sample}
\label{sec4.5.2}
Some contrastive frameworks argue that the negative samples are crucial\cite{grill2020bootstrap, he2020momentum, chen2020improved} while others say they are unnecessary\cite{simsiam}.
Hence here, we discuss the role of negative samples in GaitSSB.
Specifically, 
we remove the negative samples by simply replacing the InfoNCE with cosine similarity loss, 
where the latter aims to attract the positive pairs and do nothing for the negative pairs. 

\begin{figure}[tb]
\centering
\includegraphics[height=3.5cm]{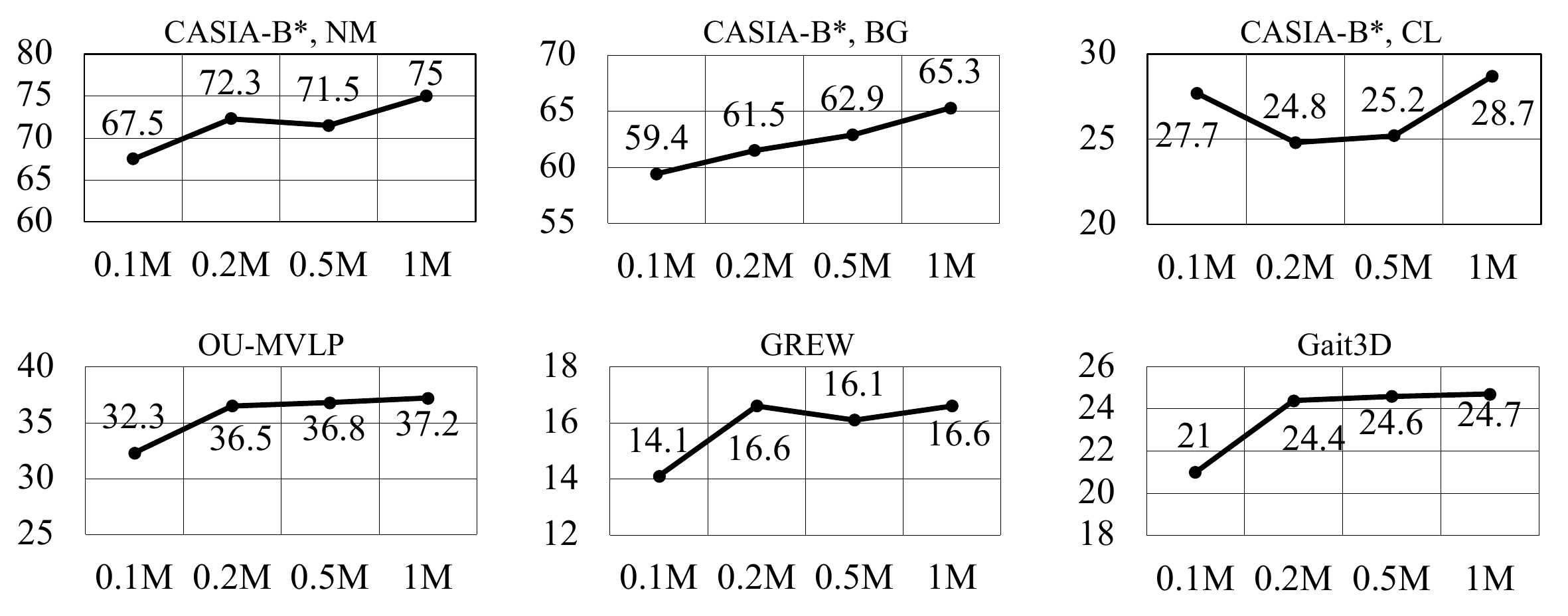}
\caption{Ablation studies: impact of the pre-training scale (w/o fine-tune).}
\label{fig7}
\vspace{-1.0em}
\end{figure}
The comparison between experiment (\textbf{a}) and (\textbf{e}) in Table~\ref{AblationStudy} declares that, 
if we remove the negative samples at the pre-training phase, 
GaitSSB would meet a prominent degeneration over all the benchmarks, 
\textit{i.e.}, CASIA-B* (NM: $-31.6\%$, BG: $-33.2\%$, CL: $-15.5\%$), OU-MVLP ($-22.6\%$), GREW($-7.5\%$) and Gait3D ($-9.2\%$).

As far as we know, 
most of the contrastive frameworks are developed on coarse-grained tasks, 
\textit{e.g.}, the image recognition and person re-identification, where a noticeable feature is that two different images from semantically near categories naturally own relatively more minor visual distance.
However, it is the similar camera angle instead of the same identity, being more likely to make two different walking sequences alike in visual appearance for gait data.
Hence, increasing the distance between the representation of negative pairs, especially those with a similar view, 
is crucial for GaitSSB to build a discriminative metric space.
The hypothesis proposed in Sec.~\ref{sec5} will offer a further methodological explanation for this issue. 

\subsubsection{Impact of Pre-training Scale}
To study the impact of the pre-training scale, 
we build the GaitLU-0.5M, 0.2M and 0.1M by randomly selecting walking sequences from GaitLU-1M,
where the GaitLU-0.5 is a subset of GaitLU-1M, GaitLU-0.2M is a subset of GaitLU-0.5M, and so on.

As shown in Fig.~\ref{fig7}, 
the unsupervised performance of our GaitSSB achieved on OU-MVLP, GREW and Gait3D mainly benefits from the larger pre-training scale.
However, 
for CASIA-B*, 
only the testings under NM and BG conditions are generally consistent with this observation, while the accuracy oscillates with the pre-training scale in the most challenging CL case.
The underlying reason may be that the unlabelled GaitLU-1M lacks the natural dress-changing factors, 
giving rise to the invalidity of enlarging the pre-training scale.
Therefore, 
forming positive pairs by effectively simulating the cloth-changing cases should be critical for further study.

\begin{figure}[tb]
\centering
\includegraphics[height=2.2cm]{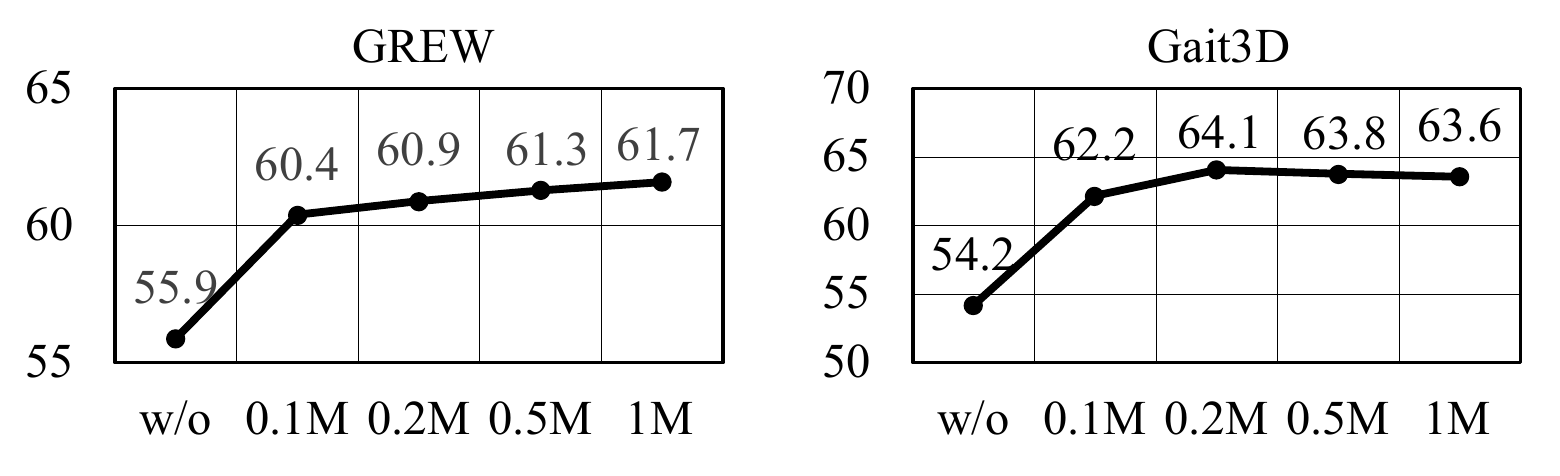}
\caption{\textcolor{black}{Ablation studies: impact of the pre-training scale (w/ fine-tune).}}
\label{pre-training_scale_transfer}
\vspace{-1.0em}
\end{figure}

\textcolor{black}{
Moreover, Fig.~\ref{pre-training_scale_transfer} presents the counterpart of Fig.~\ref{fig7} after fine-tuning the pre-trained GaitSSB on GREW and Gait3D. 
As we can see, the GaitSSB pre-trained on varying scales consistently outperforms the GaitSSB w/o pre-training, adequately showing the GaitSSB's robustness to the pre-training scale. 
Moreover, we note that the performance on GREW gradually increases with the larger pre-training scale, and the growth curve on Gait3D quickly reaches saturation. 
Taking the scale issue into consideration, \textit{i.e.}, GaitLU-1M is about 55 times larger than the training set of Gait3D and only 10 times larger than that of GREW, we think that the self-supervised data mining capacity of GaitSSB still needs to be improved to explore continual knowledge from the increasing pre-training scale. 
Therefore, further study majoring in gait data mining efficiency is highly desired.
}

\section{Hypothesis and Discussion}
\label{sec5}
Previous contrastive frameworks have provided some pioneer intuitions on the behavior of the pre-training process.
For example, 
Grill \textit{et al.} consider that the two-stream network in BYOL\cite{grill2020bootstrap} is similar to the discriminator and generator in GANs\cite{goodfellow2014generative} respectively.
Chen \textit{et al.} hypothesize that SimSiam\cite{simsiam} is an implementation of an Expectation-Maximization-like algorithm that aims at solving two underlying sub-problems.
In this section, 
we pay more attention to gait-specific issues, 
\textit{i.e.}, proposing the hypothesis on why the intra-sequence augmentation and negative samples are crucial for GaitSSB.
Also, we propose some methodological suggestions for further research.

Let a map $\Pi(\cdot)$ present the corruptions caused by various noisy factors.
For a sample's feature vector $x$ generated by our GaitSSB, 
$\Pi(x)$ denotes a set of the samples owning the identical class as $x$, 
covering a wide range of practical factors, \textit{e.g.}, carrying, dressing and camera views, \textit{etc}.
We define the intra-class and inter-class distance of sample $x$ against the noisy factors in $\Pi(\cdot)$ as:
\begin{equation}
\begin{split}
d^+_\Pi(x) &= \max \limits_{x^+\in\Pi(x)}{\left \| x - x^+ \right \|_2} \\
d^-_\Pi(x) &= \min \limits_{x^-\notin\Pi(x)}{\left \| x - x^- \right \|_2}  
\end{split}
\end{equation}
Since all the vectors are $l_2$-normalized in the obtained metric space, 
here we use the Euclidean distance $\left \| \cdot \right \|_2$ that is equivalent to the cosine similarity as a measure of the relative distance. 
Obviously, we say that the sample $x$ can be identified correctly against $\Pi(\cdot)$ if $d^+_\Pi(x)<d^-_\Pi(x)$.

For our self-supervised benchmark, 
since GaitLU-1M is unlabeled, 
we introduce the silhouette augmentation operator $\pi(\cdot)$ as a substitute to simulate noisy factors within $\Pi(\cdot)$.
Differently, 
$\pi(\cdot)$ only contains three categories of hand-designed noisy factors, 
\textit{i.e.}, 
the spatial augmentations for gait appearance changes, 
the intra-sequence augmentations for pre-treatment errors and camera angle changes, 
and the sampling augmentations for enhancing the view-invariant representation learning.
It is worth noting that the view changes inside a walking video are primarily slight, 
indicating that $\pi(\cdot)$ usually contains subtle view-changing variations.
Similarly, we define the intra and inter-class distance of sample $x$ against $\pi(\cdot)$ as $d^+_\pi(x)$ and $d^-_\pi(x)$.
Next, we concentrate on one of the most critical scenarios for gait recognition, namely the cross-view cases,
and ignore other noisy factors for simplicity.

In identical-view cases, 
there are almost no view differences between the walking videos of the probe and gallery, 
indicating $\Pi(x)\to\pi(x)$ at this time.
Therefore, the proposed GaitSSB can maintain much pre-training performance when tested on other supervised benchmarks.
The results in the diagonal of Fig.~\ref{fig6} have verified this point.

For more challenging cross-view cases, 
$\Pi(\cdot)$ covers more dramatic view-changing factors than that within $\pi(\cdot)$, 
\textit{i.e.}, $\pi(x)\subset \Pi(x)$,
intuitively meaning that it is hard for GaitSSB to maintain its pre-training performance on this occasion.
But encouragingly, 
Fig.~\ref{fig6} shows that GaitSSB can perform still promising performance in the most challenging cross-view case, 
such as $42.0\%$ rank-1 ($68.0\%$ rank-5) average accuracy for the most challenging $000^\circ$-$090^\circ$ (probe-gallery) view pair on the widely-used CASIA-B\cite{Yu2006}.
This observation encourages us to explore the further understanding of the view-invariant feature learning in our GaitSSB.
Next, we start by analyzing the inter/intra-class distance.

Since $\pi(x)\subset \Pi(x)$, 
we can 
get $\overline{\Pi}(x) \subset \overline{\pi}(x)$ where $\overline{\Pi}(x)=\left\{ x^-|x^-\notin\Pi(x)\right\}$ and $\overline{\pi}(x)=\left\{ x^-|x^-\notin\pi(x)\right\}$.
Consequently, 
\begin{equation}
\label{equ4}
d^-_\pi(x) \leqslant d^-_\Pi(x) 
\end{equation}

\noindent indicates GaitSSB can bring much of the ability to increase inter-class distance into the evaluation stage regardless of the challenging cross-view factors.
However, 
we notice that some contrastive frameworks\cite{simsiam, grill2020bootstrap} have achieved amazing performance without the design of enlarging the distance between representations of negative pairs.
We argue that the task granularity makes it different a lot: 
\begin{itemize}
    \item \textit{Data Granularity.} 
    For some coarse-grained tasks, 
    their pre-training dataset, \textit{e.g.}, the ImageNet\cite{deng2009imagenet} for image recognition, 
    usually has millions of images distributed over only thousands of classes.
    This relatively dense distribution implies that a training batch may contain two different images from the same class,
    but the contrastive model will still enlarge their distance and sequentially make a mistake.
    As for our benchmark, 
    it is almost impossible for an individual to appear twice in GaitLU-1M since the videos are collected around the world, 
    varying in locations, dates and photographers. 
    Therefore, we consider that GaitLU-1M contains only one walking sequence for each subject, 
    meaning that GaitSSB would not mistakenly repulse two different sequences owning the same identity away.
    Considering Eq.~\ref{equ4}, 
    we assume that the ability to enlarge the distance within the negative pair learned at the pre-training stage is not misleading and, thus, pretty useful for the downstream gait task.
    \item \textit{Overall Similarity.} 
    The overall visual similarity often dominates in the final representation for contrastive frameworks\cite{Wu_2018_CVPR}.
    However, for gait recognition, 
    it is a similar camera view rather than the same identity that usually makes two silhouette sequences alike in overall appearance.
    Hence, 
    using negative samples is a good choice since it tells the network needs to increase the distance within the negative pair, 
    even if two sequences are similar in the camera angle.
\end{itemize}

Combining with the comparison between experiment (\textbf{a}) with (\textbf{e}) in Table~\ref{AblationStudy}, 
we argue that learning to enlarge the distance within negative pair (using the negative samples) is crucial for contrastive gait pre-training.

\begin{figure}[tb]
\centering
\includegraphics[height=2.3cm]{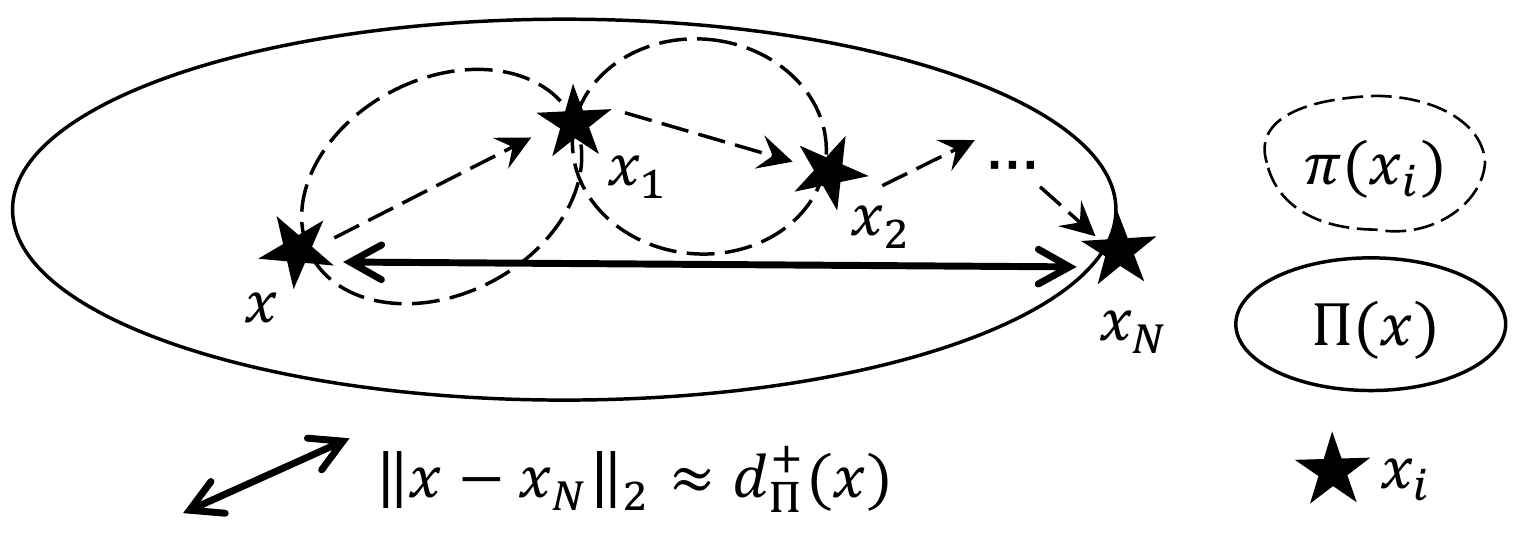}
\caption{Approximate $\Pi(x)$ by repetitively performing $x_i\in\pi(x_{i-1})$.}
\label{fig8}
\vspace{-1.0em}
\end{figure}
As to the intra-class distance,  
inspired by calculus thinking, 
we regard $\pi(\cdot)$ as the differential element of $\Pi(\cdot)$.
As shown in Fig.~\ref{fig8}, 
consider the sample $x$ can be augmented $N > 1$ times by $\pi(\cdot)$, 
\textit{i.e.}, $x_1\in\pi(x)$, $x_2\in\pi(x_1)$, ..., $x_N\in\pi(x_{N-1})$ with
\begin{equation}
\label{equ5}
x_i=\mathop{\arg\max}\limits_{x^+\in\pi(x_{i-1})}{\left\| x_{i-1} - x^+\right\|_2}
\end{equation}
where $i\in\left\{1, ...,N \right\}$, and $x_0$ here denotes $x$ (the same below).
This process goes like transferring the camera angle by a maximal offset again and again.
Then, 
we can approximate $d^+_\Pi(x)$ by finding a large enough $N$ to make:
\begin{equation}
\label{equ6}
\begin{split}
d^+_\Pi(x) &\approx \left\| x - x_N\right\|_2 \\
&\leqslant \sum_{i=1}^{N}\left\| x_{i-1} - x_i\right\|_2 \\
&=\sum_{i=1}^{N}d^+_\pi(x_i)
\end{split}
\end{equation}

\noindent Above inequality holds for Euclidean-metric space.
Obviously, 
the larger $N$ implies the larger camera angle difference between the probe ($x$) and gallery ($x_N$)
\footnote{
We consider that the view differences between the similar or symmetric camera angle pairs are relatively small, 
\textit{e.g.}, $000^\circ$ with $015^\circ$ or $180^\circ$.
The same to below.
}.
Eq.~\ref{equ6} displays the distance transitivity law in Euclidean-metric space, 
and implies that we can pull $x$ towards $x_N$ (reduce intra-distance $d^+_\Pi(x)$)
by minimizing the distance between $x_i$ and $x_{i-1}$ where $x_i\in\pi(x_{i-1})$ and $i\in\left\{1, ...,N \right\}$. 

Next, we make a general assumption that the well-pre-trained GaitSSB can ensure $d^+_\pi(x_i) \leqslant a \leqslant b \leqslant d^-_\pi(x_i)$ for sample $x_i$ against $\pi(\cdot)$ where $i\in\left\{1, ...,N \right\}$.
Taking Eq.~\ref{equ4} and \ref{equ6} into consideration, we know that:
\begin{equation}
\begin{split}
    b & \leqslant d^-_\Pi(x) \\
    d^+_\Pi(x) &\leqslant N a 
\end{split}
\label{equ7}
\end{equation}
showing that the lower bound of the inter-class distance of sample $x$ against the practical noisy factors in $\Pi(\cdot)$ is $b$ while the upper bound of the intra-class distance is $Na$.
%
%

For the convenience of analysis, here we fix $N$ first.
If GaitSSB can increase the distance between representations of negative pairs by a large margin at the pre-training stage, 
\textit{i.e.}, make $a \ll b$ until $N < \nicefrac{b}{a}$, 
we can get $d^+_\Pi(x) < d^-_\Pi(x)$ at this time.
Our self-supervised framework can rightly recognize the sample $x$ against the challenging cross-view factors in $\Pi(\cdot)$ at the supervised testing stage.
Therefore, 
increasing $b$ and meanwhile decreasing $a$ via enhancing the recognition capacity of the backbone is vital for GaitSSB.

On the other hand, the larger cross-view angle implies the larger $N$, 
showing that it is harder to ensure $N < \nicefrac{b}{a}$, especially for those cross-large-view cases.
The salient performance degradation around the diagonal/anti-diagonal of Fig.~\ref{fig6} can verify this point experimentally.
Moreover, 
the degree of view-changing transition involved by $\pi(\cdot)$, 
which is regarded as the differential element of $\Pi(\cdot)$ as shown in Fig.~\ref{fig8}, 
make it different as well.
Intuitively, 
the more dramatic view-changing factors that $\pi(\cdot)$ contain, 
the smaller $N$ we need to approximate $\Pi(\cdot)$.
If we remove the intra-sequence and sampling augmentation strategies from $\pi(\cdot)$,  
it will include almost no view-changing factors as discussed in Sec.~\ref{AblationStudy.1}.
At this time, 
it is impossible to ensure $N < \nicefrac{b}{a}$ since $N\to\infty$.
The performance collapse in Table~\ref{AblationStudy} (experiment (\textbf{a}) \textit{v.s.} (\textbf{c})) can stand for it point experimentally.
Therefore, exploring the more advanced augmentation strategy dominating view-changing variations dramatically should be vital for further research.

Last but not least, 
the most challenging bag-carrying and dress-changing cases, 
which have been neglected for clarity analysis, 
are worth being discussed seriously as well.
Intuitively, 
the spatial augmentation strategies should matter in learning the clothing-invariant representations.
However, 
though its effectiveness has been verified by experiment (\textbf{b}) in Table~\ref{AblationStudy}, 
our silhouette augmentation operator is far from effectively simulating the enormous variety of practical appearance changes.
This issue still needs further systematic explorations.

\section{Conclusion}
\label{sec6}
Since manual labeling is costly and insatiable, 
the real-world applications of gait recognition have been subject to small-scale supervised datasets for a long time.
Inspired by the development of self-supervised learning, 
this paper makes an attempt to verify the possibility of learning the generalizable and discriminative gait representations from massive unlabelled walking videos from the aspects of both experimental and theoretical analysis.
Technically, 
we propose a comprehensive and systematical benchmark, 
involving a large-scale unlabelled gait dataset GaitLU-1M,
and a structurally concise yet experimentally powerful baseline model GaitSSB.
We hope this work can serve as a stepping stone to learning the general gait representation by self-supervised approaches, 
and inspire future works to narrow the gap between theory and practice for gait recognition. 

\section{Ethical Statements}
\label{sec7}
The improper use or abuse of gait recognition will significantly threaten personal privacy. 
\textcolor{black}{
As of now, gathering web data on a large scale has become a common practice that significantly promotes AI research and applications. 
However, the collection inevitably involves personal data since most Internet data is created by and related to individuals. 
To protect personal privacy, most laws from different countries impose restrictions on the use of personal data by private businesses, while providing derogations for security and research purposes. 
For instance, according to General Data Protection Regulation passed by the European Parliament, in the version of the OJ L 119, 04.05.2016~\cite{GDPR}, personal data can be processed for scientific research purposes without demanding for the data subject's consent if adequately minimized and anonymized. 
Though this practice is generally perceived as legally acceptable, it introduces an unsolved ethical conundrum at the intersection of privacy protection and advancements in AI.
}
On this background, the distributor of the GaitLU-1M dataset would like to try the best to protect personal data by taking the following measures.

\begin{itemize}
    \item[a)] Data Minimization. The data of GaitLU-1M is exhibited as the sequence of binary silhouettes and body skeletons. No visual RGB information will be available.
    \item[b)] Expiration Date. GaitLU-1M can only be used for 20 years since the corresponding paper is published. After this date, all data will be deleted and not allowed to be used.
    \item[c)] Opt-out. For each gait sequence in GaitLU-1M, its data will be deleted once identified by some algorithms or people with providing confirmed evidence.
    \item[d)] Unattended Pre-treatment. The pre-processing of GaitLU-1M is based on automatic algorithms, meaning that nobody has manually processed personal visual data.
    \item[e)] Reasonable Anonymization. The raw data of GaitLU-1M, including video links and other intermediate results, have been deleted. In addition, weak annotations such as the date and city information have been erased by randomly numbering the sequences.
    \item[f)] Purpose Limitation. GaitLU-1M can be distributed only for non-commercial research purposes with a case-by-case access application.
    \item[g)] Ethical Requirements. Applicants are required to sign an additional agreement about data security and privacy protection. Conducts like identity inferring, manual labeling, private distribution, inversion attack, and biased analysis are prohibited explicitly. 
\end{itemize}

We hope the above measures can inspire further study to balance AI development and privacy protection.


\section*{Acknowledgment}
We would like to thank the helpful discussion with Mrs. Yunjie Peng, Dr. Chunshui Cao, Dr. Xu Liu and Dr. Hanyang Peng. 
This work was supported in part by the National Natural Science Foundation of China under Grant 61976144, Grant 62276025 and Grant 62206022, in part by the National Key Research and Development Program of China under Grant 2020AAA0140002, 
and in part by the Shenzhen Technology Plan Program (Grant No. GJHZ20220913142611021 and KQTD20170331093217368).

\bibliographystyle{IEEEtran}
\bibliography{ref}
\vspace{-5mm}
\begin{IEEEbiography}[{\includegraphics[width=1in,height=1.25in,clip,keepaspectratio]{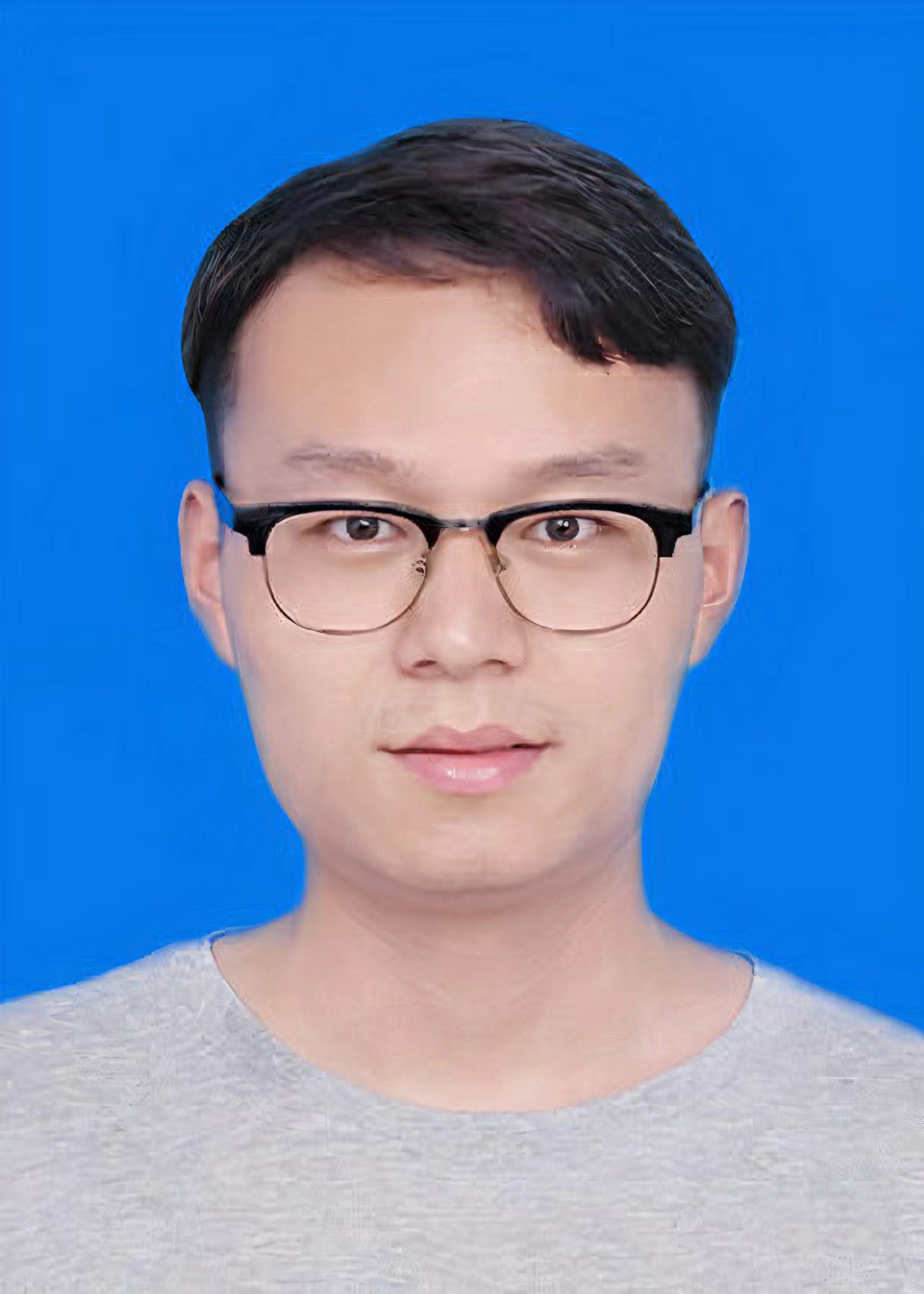}}]{Chao Fan}
 received the B.E. and M.S. degrees from Xi'an University of Technology in 2018 and University of Science and Technology Beijing in 2021, respectively.
 He is currently a Ph.D. candidate with Department of Computer Science and Engineering, Southern University of Science and Technology.
 His research interests include gait recognition and contrastive learning.
\end{IEEEbiography}

\vspace{-5mm}
\begin{IEEEbiography}[{\includegraphics[width=1in,height=1.25in,clip,keepaspectratio]{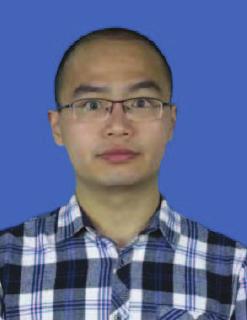}}]{Saihui Hou}
 received the B.E. and Ph.D. degrees from University of Science and Technology of China in 2014 and 2019, respectively.
 He is currently an Assistant Professor with School of Artificial Intelligence, Beijing Normal University.
 His research interests include computer vision and machine learning.
 He recently focuses on gait recognition which aims to identify different people according to the walking patterns.
\end{IEEEbiography}

\vspace{-5mm}
\begin{IEEEbiography}[{\includegraphics[width=1in,height=1.25in,clip,keepaspectratio]{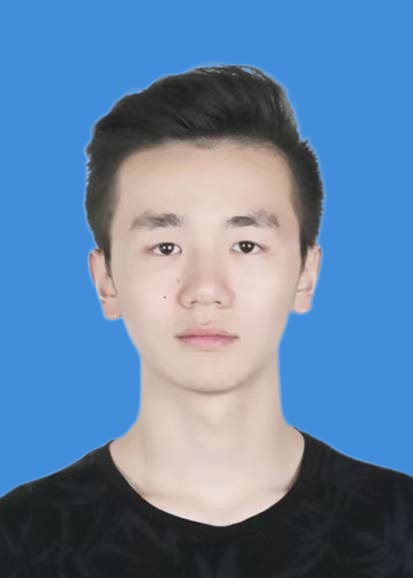}}]{Jilong Wang}
received his B.S. degree from the University of Science and Technology Beijing in 2019.
He is currently a Ph.D. student at the University of Science and Technology of China and studies in the Center for Research on Intelligent Perception and Computing (CRIPAC), National Laboratory of Pattern Recognition (NLPR), Institute of Automation, Chinese Academy of Sciences (CASIA).
His research interests are in artificial intelligence, machine learning and computer vision.
\end{IEEEbiography}

\begin{IEEEbiography}[{\includegraphics[width=1in,height=1.25in,clip,keepaspectratio]{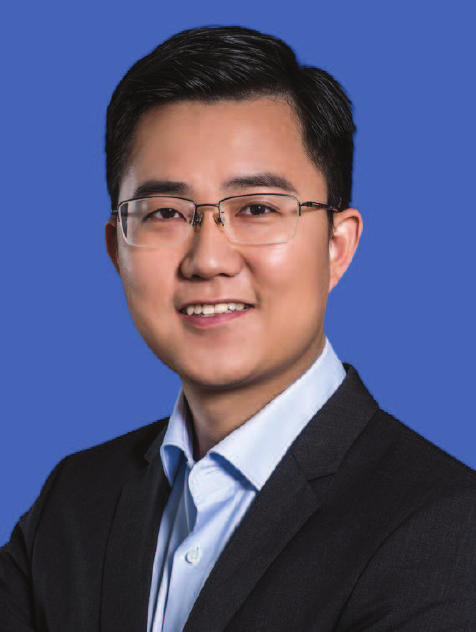}}]{Yongzhen Huang}
 received the B.E. degree from Huazhong University of Science and Technology in 2006, and the Ph.D. degree from Institute of Automation, Chinese Academy of Sciences in 2011.
 He is currently an Associate Professor with School of Artificial Intelligence, Beijing Normal University.
 He has published one book and more than 80 papers at international journals and conferences such as TPAMI, IJCV, TIP, TSMCB, TMM, TCSVT, CVPR, ICCV, ECCV, NIPS, AAAI.
 His research interests include pattern recognition, computer vision and machine learning.
\end{IEEEbiography}

\begin{IEEEbiography}[{\includegraphics[width=1in,height=1.25in,clip,keepaspectratio]{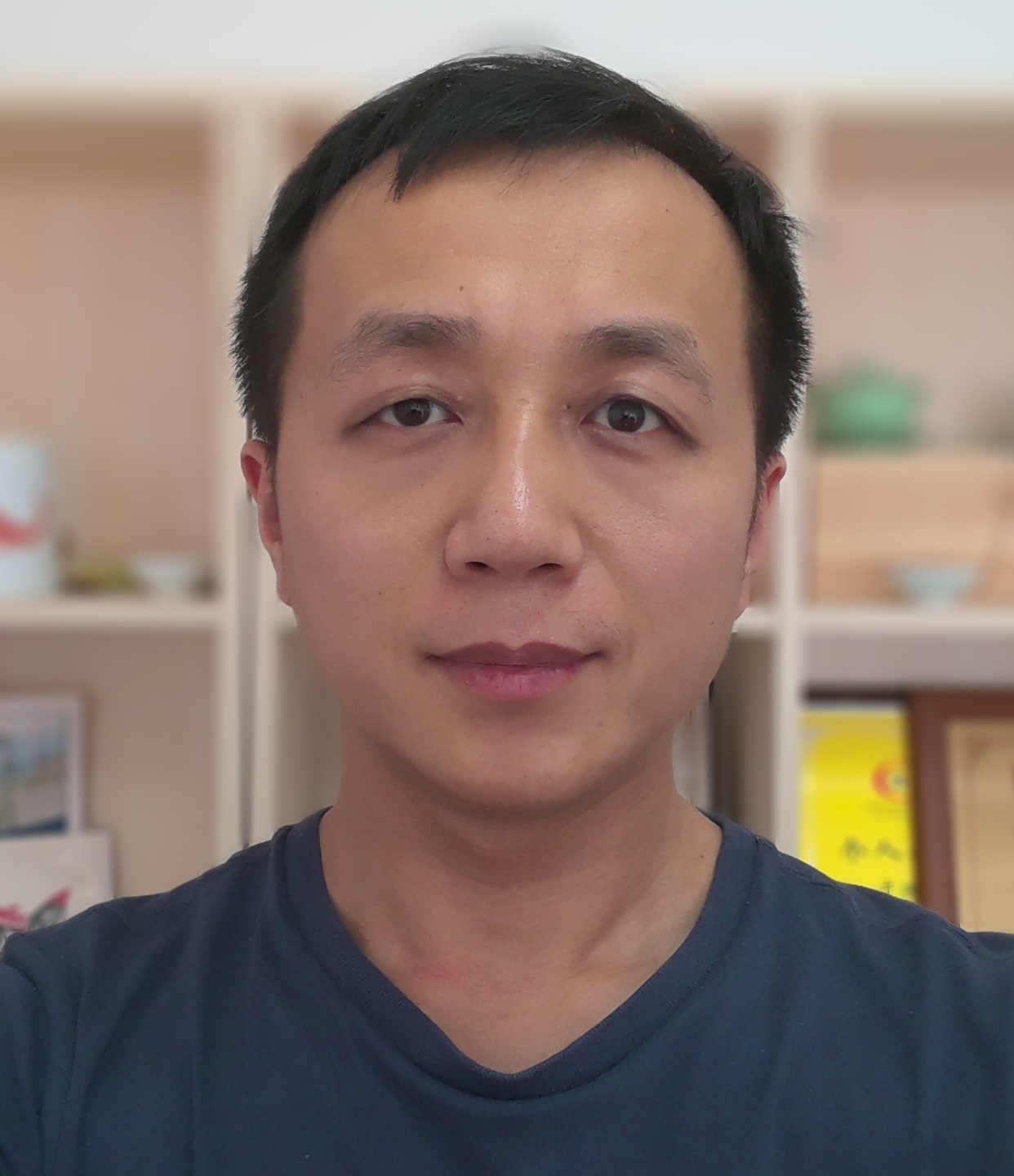}}]{Shiqi Yu} is currently an Associate Professor in the Department of Computer Science and Engineering, Southern University of Science and Technology, China. He received his B.E. degree in computer science and engineering from the Chu Kochen Honors College, Zhejiang University in 2002, and Ph.D. degree in pattern recognition and intelligent systems from the Institute of Automation, Chinese Academy of Sciences in 2007. He worked as an Assistant Professor and an Associate Professor in Shenzhen Institutes of Advanced Technology, Chinese Academy of Sciences from 2007 to 2010, and as an Associate Professor in Shenzhen University from 2010 to 2019. His research interests include gait recognition, face detection and computer vision.
\end{IEEEbiography}
\end{document}